\documentclass[final]{cvpr}

\usepackage{times}
\usepackage{epsfig}
\usepackage{graphicx}
\usepackage{amsmath}
\usepackage{amssymb}
\usepackage{stmaryrd,multirow,subfig,color,verbatim}
\usepackage{appendix}

\usepackage[pagebackref=true,breaklinks=true,colorlinks,bookmarks=false]{hyperref}

\begin{document}

%%%%%%%%% TITLE
\title{Towards Imperceptible Universal Attacks on Texture Recognition}

\author{Yingpeng Deng$^{1}$ and Lina J. Karam$^{1,2}$\\
	$^{1}$Image, Video and Usability Lab, School of ECEE, Arizona State University, Tempe, AZ, USA\\
	$^{2}$Dept. of ECE, School of Engineering, Lebanese American University, Lebanon\\
	\tt\small{\{ypdeng,karam\}@asu.edu}}

\maketitle

%%%%%%%%% ABSTRACT
\begin{abstract}
	
	Although deep neural networks (DNNs) have been shown to be susceptible to image-agnostic adversarial attacks on natural image classification problems, the effects of such attacks on DNN-based texture recognition have yet to be explored. As part of our work, we find that limiting the perturbation's $l_p$ norm in the spatial domain may not be a suitable way to restrict the perceptibility of universal adversarial perturbations for texture images. Based on the fact that human perception is affected by local visual frequency characteristics, we propose a frequency-tuned universal attack method to compute universal perturbations in the frequency domain. Our experiments indicate that our proposed method can produce less perceptible perturbations yet with a similar or higher white-box fooling rates on various DNN texture classifiers and texture datasets as compared to existing universal attack techniques. We also demonstrate that our approach can improve the attack robustness against defended models as well as the cross-dataset transferability for texture recognition problems.
	
\end{abstract}

%%%%%%%%% BODY TEXT

\section{Introduction}

\begin{figure}[tb]
	\centering
	
	\begin{minipage}{0.14\textwidth}
		\centering
		\small
		no attack\\
		\includegraphics[width=1\textwidth]{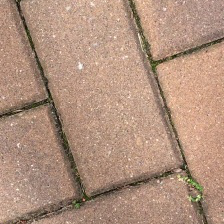}\\
		\footnotesize
		\textcolor[rgb]{0,0.5,0}{brick 100\%}\\
		\footnotesize{\ }\\
		\tiny{\ }\\
		\includegraphics[width=1\textwidth]{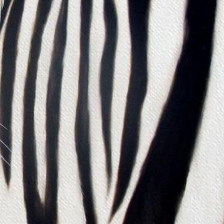}\\
		\footnotesize
		\textcolor[rgb]{0,0.5,0}{stripped 15\%}\\
		\footnotesize{\ }\\
		\tiny{\ }\\
		\includegraphics[width=1\textwidth]{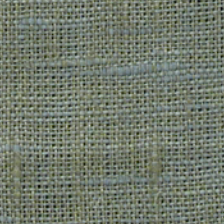}\\
		\footnotesize
		\textcolor[rgb]{0,0.5,0}{linen 54\%}\\
		\footnotesize{\ }\\
	\end{minipage}
	\begin{minipage}{0.14\textwidth}
		\centering
		\small
		GAP-tar~\cite{poursaeed2018generative}
		\includegraphics[width=1\textwidth]{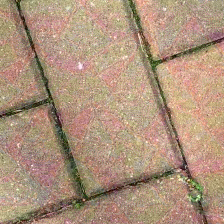}\\
		\footnotesize
		\textcolor[rgb]{0.7,0,0}{aluminum 66\%}\\
		STSIM: 0.8969\\
		\tiny{\ }\\
		\includegraphics[width=1\textwidth]{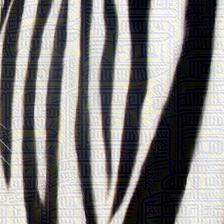}\\
		\footnotesize
		\textcolor[rgb]{0,0.5,0}{stripped 14\%}\\
		STSIM: 0.8707\\
		\tiny{\ }\\
		\includegraphics[width=1\textwidth]{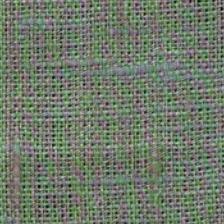}\\
		\footnotesize
		\textcolor[rgb]{0,0.5,0}{linen 47\%}\\
		STSIM: 0.9395\\
	\end{minipage}
	\begin{minipage}{0.14\textwidth}
		\centering
		\small
		FTGAP (ours)
		\includegraphics[width=1\textwidth]{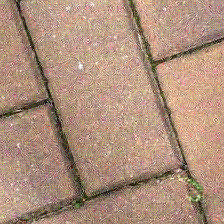}\\
		\footnotesize
		\textcolor[rgb]{0.7,0,0}{moss 100\%}\\
		STSIM: 0.9059\\
		\tiny{\ }\\
		\includegraphics[width=1\textwidth]{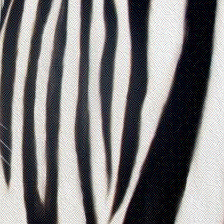}\\
		\footnotesize
		\textcolor[rgb]{0.7,0,0}{grid 5\%}\\
		STSIM: 0.9260\\
		\tiny{\ }\\
		\includegraphics[width=1\textwidth]{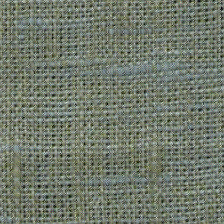}\\
		\footnotesize
		\textcolor[rgb]{0.7,0,0}{corduroy 51\%}\\
		STSIM: 0.9675\\
	\end{minipage}
	\caption{Comparison between GAP-tar~\cite{poursaeed2018generative} and our FTGAP universal attack methods. We show the visual examples of clean images (first column), GAP-tar (second column) and our FTGAP (third column) on different texture datasets (from top to bottom: GTOS~\cite{xue2017differential, xue2018deep}, DTD~\cite{cimpoi2014describing} and KTH~\cite{caputo2005class}) in each row. Under each image are the groundtruth class (green) or perturbed class (red) with the corresponding confidence scores. The similarity index STSIM~\cite{zujovic2013structural} of each perturbed image in the second \& third columns with the corresponding clean image in the first column is also given below the classification result.}
	\label{fig:viseg}
\end{figure}

While DNNs achieved considerable success in various computer vision problems~\cite{Simonyan15, szegedy2015going, he2016deep, long2015fully, ronneberger2015u, girshick2014rich, redmon2016you}, they were also shown to be vulnerable to adversarial attacks~\cite{goodfellow2014explaining, carlini2017towards, moosavi2017universal, xie2017adversarial, hendrik2017universal, li2020universal}. An adversarial attack generally produces a small perturbation which is added to the input image(s) to fool the DNN, even though such a perturbation is not enough to degrade the image quality and fool the human vision. Many existing adversarial attacks are white-box attacks, where one can have full knowledge about the targeted DNN model such as the architecture, weight and gradient information for computing the perturbation. Adversarial attacks can be categorized as image-dependent~\cite{szegedy2013intriguing, goodfellow2014explaining, moosavi2016deepfool, kurakin2016adversarial, papernot2016limitations, carlini2017towards} or image-agnostic (universal)~\cite{moosavi2017universal, poursaeed2018generative, shafahi2020universal, 9191288} attacks. Compared to the former, the latter can be more challenging to generate given that a universal perturbation is often computed to significantly lower the performance of the targeted DNN model on a whole training image set as well as on unseen validation/test sets.

Although many current universal attack methods have been designed for classification tasks involving natural images (e.g., ImageNet~\cite{ILSVRC15}), little work has been done to explore the existence of effective universal attacks against DNNs for texture image recognition tasks~\cite{zhang2017deep, xue2018deep, hu2019multi}. Thus in this paper, we explore the generation of universal adversarial attacks on texture image recognition.

Existing universal attack methods usually limit the perceptibility of the computed perturbation by setting a small and fixed $l_p$ norm threshold ($\left\| \delta \right\|_\infty \leq 10$ for example). This type of thresholding method treats different image/texture regions identically and does not take human perception sensitivity to different frequencies into consideration. In this work, we propose a frequency-tuned universal attack method, which is shown to further improve the effectiveness of the universal attack in terms of increasing the fooling rate while reducing the perceptibility of the adversarial perturbation. Instead of optimizing the perturbation directly in the spatial domain, the proposed method generates a frequency-adaptive perturbation by tuning the computed perturbation according to the frequency content. For this purpose, we adopt a perceptually motivated just noticeable difference (JND) model based on frequency sensitivity in different frequency bands. A similar method has shown its potential on the natural image classification problem~\cite{deng2020frequency}. As shown in Figure~\ref{fig:viseg}, the proposed universal attack algorithm can easily convert correct predictions into wrong labels with less perceptible perturbations.

Our contributions are summarized as follows:

\begin{itemize}\vspace*{-5pt}
	\item We examine the effectiveness of four current universal attack algorithms~\cite{moosavi2017universal, poursaeed2018generative, shafahi2020universal, 9191288} by attacking four DNN-based texture classifiers~\cite{he2016deep, zhang2017deep, xue2018deep, hu2019multi} that are trained on six different texture recognition datasets~\cite{bell2015material, xue2017differential, xue2018deep, cimpoi2014describing, wang20164d, sharan2013recognizing, caputo2005class}.\vspace*{-5pt}
	\item We propose a frequency-tuned universal attack method to generate a less perceivable perturbation for texture images while providing a similar or better performance in terms of white-box fooling rate as compared to the state-of-the-art.\vspace*{-5pt}
	\item We demonstrate that our proposed method is more robust against a universal adversarial training defense strategy~\cite{shafahi2020universal} and possesses better out-of-distribution generalization when tested across datasets.\vspace*{-5pt}
	\item With the standard data augmentation strategy~\cite{krizhevsky2012imagenet, zhang2017deep}, our proposed method can maintain a robust performance even when the training image dataset is significantly reduced in size (i.e., to $\approx$ 0.1\% of the training dataset).
\end{itemize}

\section{Related Work}

\subsection{Texture Recognition}

Texture recognition has been widely explored for many decades. Right before the prevalence of DNNs, discriminative features were commonly extracted from texture patterns through diverse types of methods, such as Scale-Invariant Feature Transform (SIFT)~\cite{lowe2004distinctive}, Bag-of-Words (BoWs)~\cite{csurka2004visual}, Vector of Locally Aggregated Descriptors (VLAD)~\cite{jegou2010aggregating}, Fisher Vector (FV)~\cite{perronnin2010improving}, etc. Cimpoi \textit{et al.}~\cite{cimpoi2015deep} used pretrained DNNs to extract deep features from texture images and achieved cutting-edge performance.

Later, research works turned to adaptive modifications of end-to-end DNN architectures. Andrearczyk and Whelan~\cite{ANDREARCZYK201663} took AlexNet~\cite{krizhevsky2012imagenet} as backbone and concatenated computed energy information with convolutional layers to the first flatten layer. Lin and Maji~\cite{lin2016visualizing} introduced the bilinear convolutional neural network~\cite{lin2015bilinear} to compute the outer product of the feature maps as the texture descriptor. By adopting ResNet~\cite{he2016deep}, Dai \textit{et al.}~\cite{dai2017fason} captured the second order information of convolutional features by the deep bilinear model and fused it with the first order information of convolutional features before the final layer, while Zhang \textit{et al.}~\cite{zhang2017deep} proposed a deep texture encoding network (DeepTEN) by inserting a residual encoding layer with learnable dictionary codewords before the decision layer. To improve on the method of Zhang \textit{et al.}~\cite{zhang2017deep}, Xue \textit{et al.}~\cite{xue2018deep} combined the global average pooling features with the encoding pooling features through a bilinear model~\cite{freeman1997learning} in their deep encoding pooling (DEP) network, followed by a multi-level texture encoding and representation (MuLTER)~\cite{hu2019multi} to aggregate multi-stage features extracted using the DEP module.

\subsection{Image-Dependent Attacks}

\begin{figure*}[t]
	\centering
	\includegraphics[width=1\textwidth]{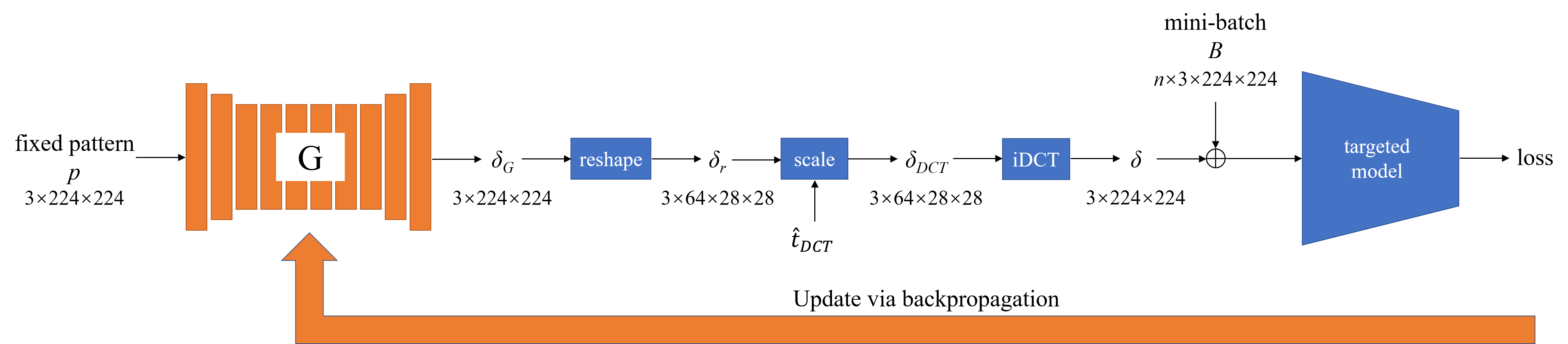}
	\caption{Block diagram of the proposed frequency-tuned generative adversarial perturbation (FTGAP). $p$ is a fixed input pattern, that is randomly sampled from a uniform distribution, of the generator $G$. For $G$, we adopt a ResNet generator, consisting of 8 consecutive residual blocks~\cite{poursaeed2018generative}. $\delta_G$ and $\delta_r$ denote the intermediate perturbations of the output of the generator and reshaping stage, respectively. $\delta_{DCT}$ is obtained after scaling using the computed DCT-domain threshold set $\hat{t}_{DCT}$, and is transformed by the iDCT into a spatial-domain perturbation $\delta$, which is added to the stochastic training batch $B$. The loss is computed to update the generator (orange) only during training via backpropagation while the other parts (blue) are kept frozen.}
	\label{fig:flow}
\end{figure*}

An image-dependent perturbation is usually computed for each input image separately to maximize the attack power. Aiming to maximize the loss between the prediction and the true label or designed logit distance function, researchers came up with various image-dependent white-box attack methods such as box-contrained L-BFGS~\cite{szegedy2013intriguing}, JSMA~\cite{papernot2016limitations}, DeepFool~\cite{moosavi2016deepfool}, C\&W~\cite{carlini2017towards} and the family of FGSM based attacks~\cite{goodfellow2014explaining, kurakin2016adversarial, dong2018boosting, xie2019improving, dong2019evading}, while constraining the $l_p$ norm of the adversarial perturbation.

However, $l_p$-norm constraints may not be suitable in limiting the perturbation perceptibility~\cite{sharif2018suitability}. In order to reduce the perturbation perceptibility for image-dependent attacks, alternative approaches were proposed to generate sparse adversarial perturbations~\cite{su2019one, xu2018structured, modas2019sparsefool, croce2019sparse, fan2020sparse}, while others were based on perceptually motivated measurements with the constraints formulated in terms of human perceptual distance~\cite{luo2018towards}, pixel-wise JND models~\cite{wang2019invisible, zhang2020advjnd} and perceptual color distance~\cite{zhao2020towards}. Overall, these image-dependent methods aimed to compute the imperceptibility metrics (either sparsely perturbed pixel locations or perceptual measurement) directly using image-specific information of the perturbed image. In many cases, even without a perceptibility constraint, image-dependent attacks can be easily made invisible by finding the minimal effective perturbation (e.g., \cite{moosavi2016deepfool}). In contrast, our proposed method generates image-content-agnostic perturbations, in which case reducing the perceptibility is more chanllenging given that the evaluated images are inaccesible during training. Furthermore, in our proposed approach, the perturbation perceptibility is constrained by adaptively tuning the perturbation according to frequency bands' characteristics.

Adversarial attacks in the frequency domain have already been explored, many of which limited or emphasized the perturbations on low frequency content to construct effective image-dependent attacks~\cite{zhou2018transferable, guo2018low, sharma2019effectiveness, dabouei2020smoothfool, shamsabadi2020colorfool} or computationally efficient black-box attacks~\cite{guo2019simple, liu2019geometry}. Yin \textit{et al.}~\cite{yin2019fourier} constructed image-dependent perturbations by randomly corrupting up to two sampled Fourier bands.

\subsection{Universal Attacks}

A universal, image-agnostic perturbation can drastically reduce the prediction accuracy of the targeted DNN model when applied to all the images in the training/validation/testing set. Based on DeepFool~\cite{moosavi2016deepfool}, Moosavi-Dezfooli \textit{et al.}~\cite{moosavi2017universal} generated universal adversarial perturbations (UAP) by accumulating the updates iteratively for all the training images. Later, some works such as generative adversarial perturbation (GAP)~\cite{poursaeed2018generative} and network for adverary generation (NAG)~\cite{reddy2018nag} adopted generative adversarial networks (GANs)~\cite{goodfellow2014generative} to produce universal perturbations. Recently, the authors of~\cite{mummadi2019defending} and~\cite{shafahi2020universal} generated universal adversarial perturbations by adopting a mini-batch based stochastic projected gradient descent (sPGD) during training and maximizing the average loss over each mini-batch. Li \textit{et al.}~\cite{li2019regional} presented a method for producing transferable universal perturbations that are resilient to defenses by homogenizing the perturbation components regionally. Most recently, Deng and Karam~\cite{9191288} proposed a universal projected gradient descent (UPGD) method by integrating a PGD attack~\cite{kurakin2016adversarial} with momentum boosting in the UAP framework~\cite{moosavi2017universal}.

There are also white-box universal attack methods~\cite{mopuri-bmvc-2017, reddy2018ask, mopuri2018generalizable, liu2019universal} which do not make use of training data. These data-free methods are unsupervised and not as strong as the aforementioned supervised ones. As for frequency-domain universal attacks, Tsuzuku and Sato~\cite{tsuzuku2019structural} came up with a black-box universal attack by characterizing the computed UAP in the Fourier domain and exploring the effective perturbations that are generated on different Fourier basis functions. This frequency-domain method is black-box, whose performance is generally weaker as compared to UAP.

\section{Proposed Frequency-Tuned Universal Attack}

For generating universal adversarial perturbations, we propose a novel perceptually motivated frequency-tuned generative adversarial perturbation (FTGAP) framework, where the perturbation is generated by computing individual perturbation components in each of the frequency bands. As illustrated in Figure~\ref{fig:flow}, the integration of an inverse discrete cosine transform (iDCT) stage during the training process forces the perturbation to be generated in the DCT frequency-based domain ($\delta_{DCT}$ just before the iDCT stage in Figure~\ref{fig:flow}) and allows the generated perturbation to be adapted to the local characteristics of each frequency band so as to achieve the best tradeoff between the attack's strength and imperceptibility.

For the generator, we adopt a ResNet generator consisting of eight consecutive residual blocks as in~\cite{poursaeed2018generative}. First, a $3 \times 224 \times 224$ (channel first) pattern is randomly sampled from a uniform distribution between $0$ and $1$ and then fixed during training. By inputing the fixed pattern into a ResNet generator ended with a tanh function, the generated output with the same size of $3 \times 224 \times 224$ has all the element values in the range of $-1$ to $1$. The resulting generator's output is reshaped into a tensor $\delta_r(c,k,\hat{n}_1,\hat{n}_2)$ of size $3 \times 64 \times 28 \times 28$, where 64 denotes the number of frequency bands (flattened $8 \times 8$ DCT block) and $28 \times 28$ indicates the number of $8 \times 8$ blocks in each $224 \times 224$ input channel. For the frequency band $k$, the reshaped output tensor $\delta_r$ is scaled to ensure that the $l_\infty$ norm of each frequency band is limited by the corresponding $\hat{t}_{DCT}$ (Section~\ref{sec:jnd}), resulting in the DCT-domain perturbation $\delta_{DCT}$. The generated DCT-domain universal perturbation $\delta_{DCT}$ is transformed by the iDCT into a spatial-domain perturbation, that is added to the stochastic mini-batch $B$. We then feed the perturbed image batch, which is clipped to the valid image scale (e.g., $[0, 255]$ for 8-bit images), into the targeted model, i.e., the DNN-based texture classifier, and compute the loss. In~\cite{poursaeed2018generative}, the authors suggested two types of loss functions, $L_{tar}$ and $L_{llc}$, for universal attacks. Let $L_{ce}$ represent the cross-entropy loss which is used to train the targeted model. The first one, $L_{tar}$, maximizes the cross-entropy loss, $L_{ce}$, based on the prediction $\hat{y}$ and the true label (target) $y_{tar}$:
\begin{equation}
\small
L_{tar} = - L_{ce}(\hat{y}, y_{tar}).
\label{losstar}
\end{equation}
The second one, $L_{llc}$, minimizes the cross-entropy between $\hat{y}$ and the least likely class  $y_{llc}$ during prediction:
\begin{equation}
\small
L_{llc} = L_{ce}(\hat{y}, y_{llc}).
\label{lossllc}
\end{equation}
In our universal attack experiments, we found that using $L_{tar}$ results in a significantly improved performance as compared to $L_{llc}$ in terms of fooling rates (Section~\ref{adv}), so we adopt $L_{tar}$ in our FTGAP method. Note that only the generator is updated via gradient backpropagation during training. Once the training process is completed, the generator can be discarded and the computed perturbation is stored for evaluation.

Further details about the computation of the DCT/iDCT and the perceptually motivated frequency-adaptive thresholds $\hat{t}_{DCT}$, are provided in Sections~\ref{sec:dct} and~\ref{sec:jnd}, respectively.

\subsection{Discrete Cosine Transform}
\label{sec:dct}

The Discrete Cosine Transform (DCT) is a fundamental spatial-frequency transform tool which is extensively used in signal, image and video processing, especially for data compression, given its properties of energy compaction and of being a real-valued transform. We adopt the orthogonal Type-II DCT, whose formula is identical to its inverse DCT (iDCT). The DCT formula can be expressed as:
\begin{equation}
\small
X(k_1,k_2)=\sum_{n_1=0}^{N_1-1}\sum_{n_2=0}^{N_2-1}x(n_1,n_2)c_1(n_1,k_1)c_2(n_2,k_2),
\label{eqdct1}
\end{equation}
\begin{equation}
\small
c_i(n_i,k_i)=\tilde{c_i}(k_i)\cos\left(\frac{\pi(2n_i+1)k_i}{2N_i}\right),
\label{eqdct2}
\end{equation}
\begin{equation}
\small
\tilde{c_i}(k_i)=\left\{
\begin{array}{rcl}
\sqrt{\frac{1}{N_i}}, & k_i = 0\\
\sqrt{\frac{2}{N_i}}, & k_i \neq 0
\end{array}
\right.,
\label{eqdct3}
\end{equation}
where $ \ 0\leq n_i, k_i\leq N_i-1$ and $\ i = 1, 2$. In Equation~\ref{eqdct1}, $x(n_1,n_2)$ is the image pixel value at location $(n_1,n_2)$ and $X(k_1,k_2)$ is the DCT of the $N_1\times N_2$ image block (e.g., $N_1 = N_2 = 8$). Using the DCT, an $8\times 8$ spatial-domain block $x(n_1,n_2)$ can be converted to an $8\times 8$ frequency-domain block $X(k_1,k_2)$, with a total of 64 frequency bands. Given a $224\times 224$ color image, we can divide each image channel $c$ into $28\times 28 =784$ non-overlapping $8\times 8$ blocks, leading to its DCT $X_c(\hat{n}_1, \hat{n}_2, k_1, k_2)$, where $k_1, k_2$ denote 2-D frequency indices and the 2-D block indices $(\hat{n}_1, \hat{n}_2)$ satisfy $0\leq \hat{n}_1, \hat{n}_2 \leq 27$.

\subsection{Perceptual JND Thresholds}
\label{sec:jnd}

The Just Noticeable Difference (JND) is the minimum difference amount to produce a noticeable variation for human vision. Inspired by human contrast sensitivity, we adopt luminance-model-based JND thresholds in the DCT domain to adaptively constrain the $l_\infty$ norm of perturbations in different frequency bands, which can be computed as~\cite{ahumada1992luminance, hontsch2002adaptive}
\begin{equation}
\small
t_{DCT}(k_1,k_2) = \frac{MT(k_1,k_2)}{2\tilde{c_1}(k_1)\tilde{c_2}(k_2)(L_{max}-L_{min})},
\label{eqjnd}
\end{equation}
where $L_{min}$ and $L_{max}$ are the minimum and maximum display luminance, and $M = 255$ for 8-bit image. To compute the background luminance-adjusted contrast sensitivity $T(k_1,k_2)$, Ahumada and Peterson proposed an approximating parametric model based on frequency, luminance and viewing distance~\cite{ahumada1992luminance}. Usually the viewing distance is set to a constant (60 cm for example), and we use the luminance corresponding to $2^{b-1}$ if the bit depth is $b$ bits for the considered image (refer to Appendix~\ref{sec:compjnd} for computation details). The resulting DCT-based $l_\infty$ norm JND thresholds $t_{DCT}(k_1, k_2)$ are given below for 64 frequency bands ($0 \leq k_1, k_2 \leq 7$ with the top left corner corresponding to $k_1=0$ and $k_2=0$):
\begin{table}[h]
	\scriptsize
	\centering
	\label{tdct}
	\resizebox{0.45\textwidth}{!}{
	\begin{tabular}{cccccccc}
		17.31&12.24&4.20&3.91&4.76&6.39&8.91&12.60\\
		12.24&6.23&3.46&3.16&3.72&4.88&6.70&9.35\\
		4.20&3.46&3.89&4.11&4.75&5.98&7.90&10.72\\
		3.91&3.16&4.11&5.13&6.25&7.76&9.96&13.10\\
		4.76&3.72&4.75&6.25&8.01&10.14&12.88&16.58\\
		6.39&4.88&5.98&7.76&10.14&13.05&16.64&21.22\\
		8.91&6.70&7.90&9.96&12.88&16.64&21.30&27.10\\
		12.60&9.35&10.72&13.10&16.58&21.22&27.10&34.39\\
	\end{tabular}}.
\end{table}

Rather than limiting the perceptibility to be at the JND level, one can adjust/control the tolerance beyond the JND level with a coefficient matrix $\lambda(k_1, k_2)$. In this way, the final threshold for each DCT frequency band $(k_1, k_2), 0 \leq k_1, k_2 \leq 7$, can be expressed as
\begin{equation}
\small
\hat{t}_{DCT}(k_1, k_2) = \lambda(k_1, k_2) \cdot t_{DCT}(k_1, k_2).
\label{thres}
\end{equation}

Generally, a human is more sensitive to intensity changes that occur in the image regions that are dominated by low frequency content than those with high frequency content. In practice, we found that relaxing high frequency components beyond the JND thresholds can increase the effectiveness without severely weakening the imperceptibility of the perturbation. So, in order to provide a good performance tradeoff between effectiveness and imperceptibility, we compute $\lambda(k_1, k_2)$ based on the radial frequency $f(k_1,k_2)$ (see Equation~\ref{eqf} in Appendix~\ref{sec:compjnd} for the computation of the radial frequency) that is associated with each DCT frequency band $(k_1,k_2)$ as follows:
\begin{equation}
\small
\lambda(k_1,k_2) = \lambda_{l} (1-S(f(k_1,k_2))) + \lambda_{h} S(f(k_1,k_2)),
\label{lambda}
\end{equation}
where $\lambda_{l}$ ($\lambda_{h}$) is a constant parameter indicating the gain for low (high) frequency. $S(f(k_1, k_2))$ is a shifted sigmoid function about $f(k_1,k_2)$:
\begin{equation}
\small
S(f(k_1,k_2)) = \frac{1}{1 + e^{-(f(k_1,k_2)-f_{c})}},
\label{sigmoid}
\end{equation}
where $f_c$ is the cut-off frequency between low and high frequency bands. Given that $S(f(k_1,k_2))$ approaches zero when $f(k_1,k_2)\ll f_c$ and one when $f(k_1,k_2)\gg f_c$, $\lambda_{l}$ and $\lambda_{h}$ can be used to control the perceptibility in low ($f(k_1,k_2)<f_c$) and high ($f(k_1,k_2)>f_c$) frequency bands.
 
\section{Experimental Results}

\begin{table*}
	\centering
	\caption{Fooling rates by the universal attacks against different models on various texture datasets. Bold number indicates best performance and underlined number denotes the second best performance on the corresponding model in each column within the same dataset.}
	\label{fr}
	
	\resizebox{0.7\textwidth}{!}{
	\begin{tabular}[t]{|c|cccc|c|cccc|c|}
		\hline
		&ResNet&DeepTEN&DEP&MuLTER&mean&ResNet&DeepTEN&DEP&MuLTER&mean\\
		\cline{2-11}
		&\multicolumn{5}{c|}{MINC}&\multicolumn{5}{c|}{GTOS}\\
		\hline
		UAP&86.6&73.1&59.8&66.4&71.5&51.0&27.7&31.6&39.0&37.3\\
		GAP-llc&74.9&84.4&87.1&80.3&81.7&45.9&49.6&57.6&54.6&51.9\\
		GAP-tar&\textbf{94.1}&\textbf{94.8}&\textbf{94.5}&\underline{94.2}&\textbf{94.4}&69.5&\textbf{81.5}&\underline{72.5}&\textbf{79.0}&\underline{75.6}\\
		sPGD&\underline{93.8}&\underline{94.3}&93.1&93.4&93.7&61.5&74.9&70.0&76.3&70.7\\
		UPGD&93.4&93.7&93.1&93.7&93.5&\textbf{78.0}&\underline{77.5}&72.4&\underline{77.8}&\textbf{76.4}\\
		FTGAP&93.6&93.4&\underline{94.3}&\textbf{94.6}&\underline{94.0}&\underline{73.0}&70.1&\textbf{75.2}&\textbf{79.0}&74.3\\
		\hline
		&\multicolumn{5}{c|}{DTD}&\multicolumn{5}{c|}{4DLF}\\
		\hline
		UAP&26.2&36.6&32.4&38.9&33.5&83.6&72.5&79.2&77.2&78.1\\
		GAP-llc&55.7&75.3&61.4&76.0&67.1&69.7&75.6&81.1&72.2&74.7\\
		GAP-tar&\underline{72.5}&\underline{86.2}&\underline{81.1}&83.6&\underline{80.9}&87.5&84.2&\textbf{90.0}&\textbf{89.4}&\underline{87.8}\\
		sPGD&70.9&83.9&78.8&\underline{84.5}&79.5&86.4&84.4&85.6&81.7&84.5\\
		UPGD&71.8&82.6&80.5&79.9&78.7&\underline{88.3}&\underline{88.3}&\underline{86.1}&81.9&86.2\\
		FTGAP&\textbf{78.4}&\textbf{86.6}&\textbf{88.8}&\textbf{88.6}&\textbf{85.6}&\textbf{89.2}&\textbf{88.9}&\textbf{90.0}&\textbf{90.0}&\textbf{89.5}\\
		\hline
		&\multicolumn{5}{c|}{FMD}&\multicolumn{5}{c|}{KTH}\\
		\hline
		UAP&65.0&49.0&39.0&50.0&50.8&75.8&41.9&62.0&42.0&55.4\\
		GAP-llc&51.0&67.0&78.0&55.0&62.8&48.2&57.6&8.4&45.1&39.8\\
		GAP-tar&\underline{90.0}&\textbf{94.0}&\underline{88.0}&\underline{88.0}&\underline{90.0}&68.9&79.2&\underline{75.9}&80.2&76.1\\
		sPGD&89.0&\underline{93.0}&87.0&\textbf{89.0}&89.5&69.2&\underline{79.9}&72.8&\textbf{80.6}&75.6\\
		UPGD&79.0&75.0&86.0&69.0&77.3&\textbf{85.7}&78.5&74.3&75.2&\underline{78.4}\\
		FTGAP&\textbf{92.0}&\textbf{94.0}&\textbf{91.0}&86.0&\textbf{90.8}&\underline{73.6}&\textbf{82.1}&\textbf{76.2}&\textbf{85.6}&\textbf{79.4}\\
		\hline
	\end{tabular}}
\end{table*}

\textbf{Datasets and models.} We train each of the following four DNN models, ResNet~\cite{he2016deep}, DeepTEN~\cite{zhang2017deep}, DEP~\cite{xue2018deep} and MuLTER~\cite{hu2019multi} on six texture datasets, Material in Context (MINC)~\cite{bell2015material}, Ground Terrain in Outdoor Scenes (GTOS)~\cite{xue2017differential, xue2018deep}, Describable Texture Database (DTD)~\cite{cimpoi2014describing}, the 4D light-field (4DLF) material dataset, Flickr Material Dataset (FMD)~\cite{sharan2013recognizing} and KTH-TIPS-2b (KTH)~\cite{caputo2005class}, and we obtain 24 trained classifiers. We then conduct universal attacks on the 24 trained classifiers. For each of the considered dataset, universal perturbations are computed by using a subset of images that are randomly sampled from the training set. The generated perturbations are then evaluated on the testing data. The dataset descriptions and DNN model training strategies are presented in Appendix~\ref{sec:imple}. All our results are reported on the testing set of the corresponding dataset.

\textbf{Attack strategies.} Generally, no data augmentation is used while computing the universal perturbations. But in our implementation, given the small number of data samples for 4DLF, FMD and KTH, we use data augmentation to prevent overfitting. We perform universal attacks using the proposed FTGAP on the trained models. For comparison, we also conduct attacks using  UAP~\cite{moosavi2017universal}, GAP~\cite{poursaeed2018generative}, sPGD~\cite{mummadi2019defending, shafahi2020universal} and UPGD~\cite{9191288}. For GAP~\cite{poursaeed2018generative}, we refer to the perturbations computed using the loss in Equations~\ref{losstar} and~\ref{lossllc} as GAP-tar and GAP-llc, respectively. For GAP and sPGD, we use a mini-batch size of 32 when computing the perturbation. The default parameters as provided by the respective authors are used for all the compared attack methods, except for UPGD whose hyperparameters (i.e., initial learning rate and decay factor, momentum, etc.) were varied to maximize the performance for the computed perturbations in terms of fooling rate. For our FTGAP method, we use the Adam optimizer~\cite{kingma2014adam} with $\alpha=0.0002$ and $\beta_1=0.5$. We stop the training process of the perturbation and report the results when the fooling rates are unable to increase by more than 0.5\% within 5 epochs for all the attack methods, except for UAP. For UAP, we choose its best result in terms of fooling rate within 20 epochs given that its learning curves were observed to fluctuate drastically. All the compared perturbations which are computed in the spatial domain are constrained within an $l_\infty$ norm of 10 on 8-bit images ($\approx$ 0.04 for the normalized image scale $[0, 1]$). For the proposed FTGAP method, the values of the parameters $\lambda_{l}$, $\lambda_{h}$ and $f_c$ in Equations~\ref{lambda} and~\ref{sigmoid} are provided for each dataset in Appendix~\ref{sec:imple}.

\textbf{Metrics.} We adopt the top-1 accuracy and fooling rate as the performance metrics. The fooling rate describes the percentage of misclassified samples in the perturbed image set. Given an image set $I$ and a universal perturbation $\delta$ in the spatial domain, the fooling rate can be expressed as $FR=\frac{1}{n}\sum_{i=1}^{n}\llbracket\hat{y}(x_i+\delta)\neq\hat{y}(x_i)\rrbracket$, where $\llbracket\cdot\rrbracket$ denotes the indicator function, $x_i$ is the $i$-th input sample in the $n$-sample testing set and $\hat{y}(x_i)$ is the predicted label for the input $x_i$ by the classifier.

\subsection{Adversarial Strength}
\label{adv}

\begin{figure*}[!tb]
	\centering
	\begin{minipage}{0.11\textwidth}
		\centering
		\small
		MINC~\cite{bell2015material}\\
		\&\\
		DeepTEN~\cite{zhang2017deep}\\ \ \\ \ \\ \ \\ \ \\ \ \\ \ \\
		GTOS~\cite{xue2017differential, xue2018deep}\\
		\&\\
		ResNet~\cite{he2016deep}\\ \ \\ \ \\ \ \\ \ \\ \ \\ \tiny{\ \\}
		\small
		DTD~\cite{cimpoi2014describing}\\
		\&\\
		MuLTER~\cite{hu2019multi}\\ \ \\ \ \\ \ \\ \ \\ \ \\ \ \\
		4DLF~\cite{wang20164d}\\
		\&\\
		DEP~\cite{xue2018deep}\\ \ \\ \ \\ \ \\ \ \\ \ \\ \tiny{\ \\}
		\small
		FMD~\cite{sharan2013recognizing}\\
		\&\\
		ResNet~\cite{he2016deep}\\ \ \\ \ \\ \ \\ \ \\ \ \\ \ \\
		KTH~\cite{caputo2005class}\\
		\&\\
		MuLTER~\cite{hu2019multi}
	\end{minipage}
	\begin{minipage}{0.16\textwidth}
		\centering
		no attack
		\includegraphics[width=1\textwidth]{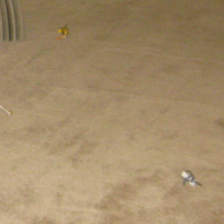}\\	
		\footnotesize{\ }\\
		\tiny{\ }\\
		\includegraphics[width=1\textwidth]{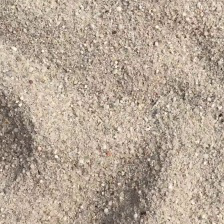}\\	
		\footnotesize{\ }\\
		\tiny{\ }\\
		\includegraphics[width=1\textwidth]{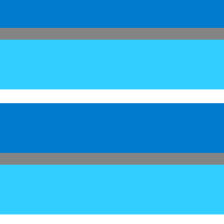}\\		
		\footnotesize{\ }\\
		\tiny{\ }\\
		\includegraphics[width=1\textwidth]{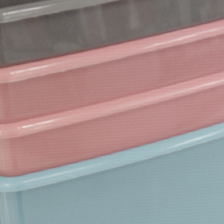}\\		
		\footnotesize{\ }\\
		\tiny{\ }\\
		\includegraphics[width=1\textwidth]{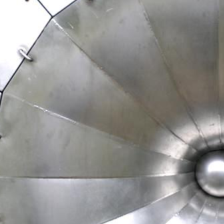}\\		
		\footnotesize{\ }\\
		\tiny{\ }\\
		\includegraphics[width=1\textwidth]{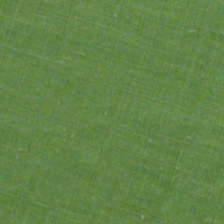}\\
		\footnotesize{\ }\\
	\end{minipage}
	\begin{minipage}{0.16\textwidth}
		\centering
		GAP-tar~\cite{poursaeed2018generative}
		\includegraphics[width=1\textwidth]{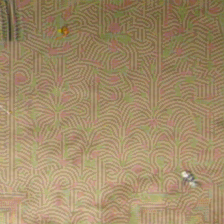}\\	
		\footnotesize{STSIM: 0.8058}\\
		\tiny{\ }\\
		\includegraphics[width=1\textwidth]{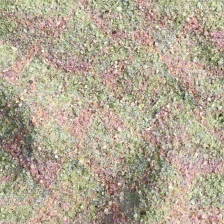}\\	
		\footnotesize{STSIM: 0.9294}\\
		\tiny{\ }\\
		\includegraphics[width=1\textwidth]{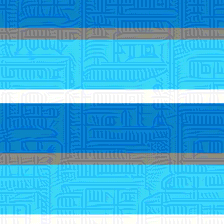}\\		
		\footnotesize{STSIM: 0.6396}\\
		\tiny{\ }\\
		\includegraphics[width=1\textwidth]{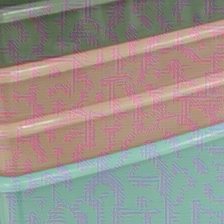}\\		
		\footnotesize{STSIM: 0.8163}\\
		\tiny{\ }\\
		\includegraphics[width=1\textwidth]{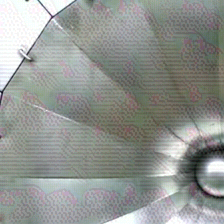}\\		
		\footnotesize{STSIM: 0.8498}\\
		\tiny{\ }\\
		\includegraphics[width=1\textwidth]{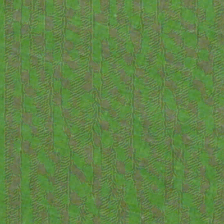}\\
		\footnotesize{STSIM: 0.8326}\\
	\end{minipage}
	\begin{minipage}{0.16\textwidth}
		\centering
		sPGD~\cite{mummadi2019defending, shafahi2020universal}
		\includegraphics[width=1\textwidth]{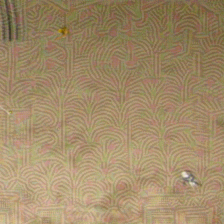}\\	
		\footnotesize{STSIM: 0.8134}\\
		\tiny{\ }\\
		\includegraphics[width=1\textwidth]{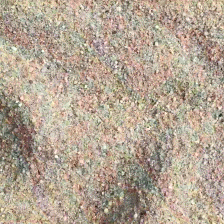}\\	
		\footnotesize{STSIM: 0.9206}\\
		\tiny{\ }\\
		\includegraphics[width=1\textwidth]{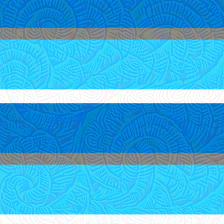}\\		
		\footnotesize{STSIM: 0.6475}\\
		\tiny{\ }\\
		\includegraphics[width=1\textwidth]{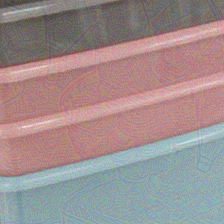}\\		
		\footnotesize{STSIM: 0.8157}\\
		\tiny{\ }\\
		\includegraphics[width=1\textwidth]{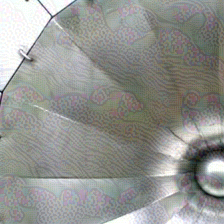}\\		
		\footnotesize{STSIM: 0.8269}\\
		\tiny{\ }\\
		\includegraphics[width=1\textwidth]{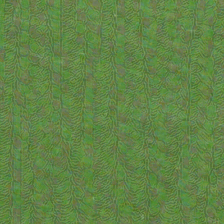}\\
		\footnotesize{STSIM: 0.8098}\\
	\end{minipage}
	\begin{minipage}{0.16\textwidth}
		\centering
		UPGD~\cite{9191288}
		\includegraphics[width=1\textwidth]{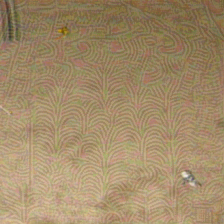}\\	
		\footnotesize{STSIM: 0.8090}\\
		\tiny{\ }\\
		\includegraphics[width=1\textwidth]{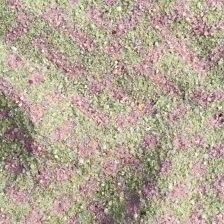}\\	
		\footnotesize{STSIM: \textbf{0.9490}}\\
		\tiny{\ }\\
		\includegraphics[width=1\textwidth]{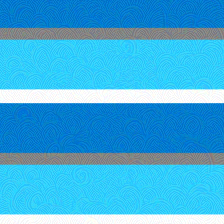}\\		
		\footnotesize{STSIM: 0.6765}\\
		\tiny{\ }\\
		\includegraphics[width=1\textwidth]{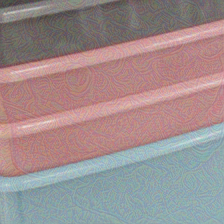}\\		
		\footnotesize{STSIM: 0.8153}\\
		\tiny{\ }\\
		\includegraphics[width=1\textwidth]{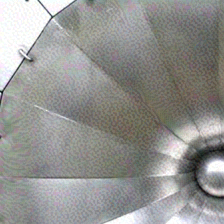}\\		
		\footnotesize{STSIM: 0.8411}\\
		\tiny{\ }\\
		\includegraphics[width=1\textwidth]{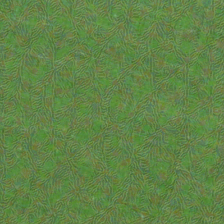}\\
		\footnotesize{STSIM: 0.8235}\\
	\end{minipage}
	\begin{minipage}{0.16\textwidth}
		\centering
		FTGAP (ours)
		\includegraphics[width=1\textwidth]{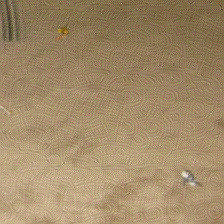}\\	
		\footnotesize{STSIM: \textbf{0.8257}}\\
		\tiny{\ }\\
		\includegraphics[width=1\textwidth]{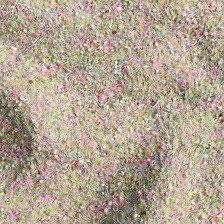}\\	
		\footnotesize{STSIM: 0.9371}\\
		\tiny{\ }\\
		\includegraphics[width=1\textwidth]{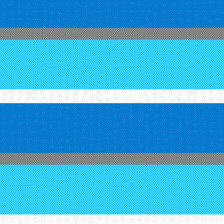}\\		
		\footnotesize{STSIM: \textbf{0.6836}}\\
		\tiny{\ }\\
		\includegraphics[width=1\textwidth]{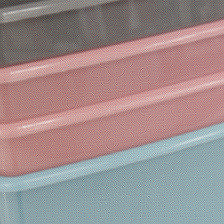}\\		
		\footnotesize{STSIM: \textbf{0.8530}}\\
		\tiny{\ }\\
		\includegraphics[width=1\textwidth]{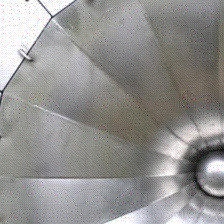}\\		
		\footnotesize{STSIM: \textbf{0.8581}}\\
		\tiny{\ }\\
		\includegraphics[width=1\textwidth]{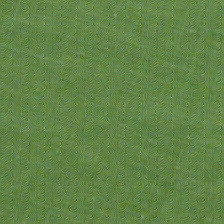}\\
		\footnotesize{STSIM: \textbf{0.8805}}\\
	\end{minipage}
	\caption{Visual comparison of different universal attack methods. Each row shows the perturbed images that result from applying on the clean image (leftmost image), universal perturbations that are generated by the considered universal attack methods for different DNN models and datasets (shown in leftmost column). The structural texture similarity (STSIM) index is given below each perturbed image. Bold number indicates best performance for the model \& dataset in each row.}
	\label{fig:viscomp}
\end{figure*}

We present the white-box fooling rates of evaluated universal attacks on the four DNN texture classifiers for all the six texture datasets in Table~\ref{fr}. Generally, GAP-tar, sPGD, UPGD and our FTGAP can perform effective universal attacks in terms of more than 70\% of fooling rates on all the datasets, showing that the DNN models are still vulnerable to universal attacks even when a small number of data samples is used for computing the perturbations in the scenario of texture recognition. From Table~\ref{fr}, it can be seen that our proposed FTGAP outperforms all other methods on DTD, 4DLF, FMD and KTH in terms of producing a higher fooling rate. For the MINC and GTOS dataset, our FTGAP method generally results in fooling rates that are comparable to GAP-tar with marginal drops of 0.4\% and 1.3\%, respectively, in terms of mean fooling rate over four DNN models.

\subsection{Perturbation Perceptibility}

Given the relatively weak performances of UAP and GAP-llc according to Table~\ref{fr}, we provide visual results only for GAP-tar, sPGD, UPGD and our FTGAP, as shown in Figure~\ref{fig:viscomp}. We show perturbed images that result from these four attack methods as well as the clean images as references, for different combinations of datasets and DNN models. Also, in order to objectively measure the perceptibility of the adversarial perturbations in the attacked images, we use the structural texture similarity (STSIM) index~\cite{zujovic2013structural}, which was shown to be a more suitable similarity index for texture content as compared to the well-known structural similarity (SSIM) index~\cite{wang2004image}.

From Figure~\ref{fig:viscomp}, it can be seen that our proposed FTGAP can produce significantly less perceptible perturbations as compared to the existing universal attack methods. In terms of the objective similary metric STSIM, our FTGAP results in the highest STSIM values for all the considered datasets and models except for the GTOS dataset with the ResNet model (2nd row in Figure~\ref{fig:viscomp}). For this latter case, the proposed FTGAP achieves the second highest STSIM value with UPGD~\cite{9191288} attaining the highest STSIM value. However, in this case, Figure~\ref{fig:viscomp} clearly shows that our proposed FTGAP method results in a significantly less perceptible perturbation as compared to all other methods, and that the STSIM is not able to accurately account for the visible color distortions that appear in the perturbed images which are generated by the existing attack methods.

It can also be shown that the performance of existing universal attack methods can be significantly improved when integrated within our proposed frequency-tuned attack framework (see Appendix~\ref{sec:alt}).

\subsection{Out-of-Distribution Generalization}

\begin{table}[t]
	\centering
	\caption{Mean generalization results over the DNN models across datasets for GAP-tar and our FTGAP. The top row describes the tested datasets and the leftmost column lists the datasets used to compute the universal adversarial perturbations. Bold number indicates best result.}
	\label{ood}
	
	\resizebox{0.47\textwidth}{!}{
		\begin{tabular}{|c|c|cccccc|c|}
			\hline
			\multicolumn{2}{|c|}{ }&MINC&GTOS&DTD&4DLF&FMD&KTH&mean\\
			\hline
			\multirow{2}*{MINC}&GAP-tar&94.4&26.8&50.6&83.0&63.8&40.3&59.8\\
			~&FTGAP&94.0&18.9&73.6&81.3&75.3&33.1&\textbf{62.7}\\
			\hline
			\multirow{2}*{GTOS}&GAP-tar&55.3&75.6&37.4&79.1&47.3&47.2&57.0\\
			~&FTGAP&42.0&74.3&74.1&86.6&76.5&66.3&\textbf{70.0}\\
			\hline
			\multirow{2}*{DTD}&GAP-tar&39.0&10.8&80.8&78.6&58.0&19.1&47.7\\
			~&FTGAP&26.1&9.3&85.6&81.7&72.5&19.4&\textbf{49.1}\\
			\hline
			\multirow{2}*{4DLF}&GAP-tar&38.8&24.7&40.3&87.8&46.0&25.9&43.9\\
			~&FTGAP&30.9&12.0&71.0&89.5&72.8&28.5&\textbf{50.8}\\
			\hline
			\multirow{2}*{FMD}&GAP-tar&51.5&16.9&49.3&80.8&90.0&20.4&51.5\\
			~&FTGAP&29.2&11.8&75.6&84.9&90.8&27.3&\textbf{53.3}\\
			\hline
			\multirow{2}*{KTH}&GAP-tar&48.8&35.0&40.1&79.1&50.3&76.1&\textbf{54.9}\\
			~&FTGAP&27.3&19.4&61.5&74.9&61.3&79.4&54.0\\
			\hline
			\multirow{2}*{mean}&GAP-tar&54.6&31.6&49.8&81.4&59.2&38.2&52.5\\
			~&FTGAP&41.6&24.3&73.6&83.2&74.8&42.3&\textbf{56.6}\\
			\hline
	\end{tabular}}
\end{table}

To test the out-of-distribution generalization of the perturbations,  we evaluate the cross-dataset performance over different datasets for each DNN model and report the average fooling rates over all the considered four DNN models on each dataset for GAP-tar and our FTGAP in Table~\ref{ood}. For the perturbations computed on each dataset (listed in the leftmost column), Table~\ref{ood} shows that the proposed FTGAP method exhibits a better out-of-distribution generalization in terms of achieving higher mean fooling rates over all the tested datasets (listed in the top row) for all the generated universal perturbations except for the one that is generated using the KTH dataset for which FTGAP achieves a mean fooling rate that is comparable to GAP-tar.

\subsection{Robustness against Defended Models}

\begin{table}[t]
	\centering
	\caption{Mean top-1 accuracy results over the defended DNN models. Bold number indicates best performance and underlined number denotes second best performance for each dataset.}
	\label{robust}
	
	\resizebox{0.47\textwidth}{!}{
		\begin{tabular}{|c|c|cccccc|}
			\hline
			~&no attack&UAP&GAP-llc&GAP-tar&sPGD&UPGD&FTGAP\\
			\hline
			MINC&78.6&76.5&63.0&\underline{59.5}&70.7&68.4&\textbf{39.0}\\
			GTOS&73.3&74.1&70.1&66.1&70.0&\underline{65.5}&\textbf{64.8}\\
			DTD&64.8&63.9&48.8&46.1&\underline{46.0}&48.8&\textbf{38.0}\\
			4DLF&64.6&56.7&43.8&\underline{36.0}&47.9&50.7&\textbf{26.9}\\
			FMD&71.3&63.8&55.8&\underline{50.3}&52.0&59.0&\textbf{45.0}\\
			KTH&78.8&75.1&72.2&\textbf{47.5}&\underline{50.4}&62.4&50.5\\
			\hline
			mean&71.9&68.4&59.0&\underline{50.9}&56.2&59.1&\textbf{44.0}\\
			\hline
	\end{tabular}}
\end{table}

We also provide comparisons of perturbation robustness against defended models in Table~\ref{robust}. For this purpose, we retrain the four considered DNN models by adopting the universal adversarial training strategy, a simple yet effective adversarial training method to defend against universal attacks~\cite{shafahi2020universal}. All the perturbations are computed on the undefended models and then applied to the input images of the defended models. From the results in Table~\ref{robust}, our FTGAP method yields the most robust perturbations as compared to the existing universal attack methods by achieving the lowest mean top-1 accuracy over the defended models for almost all the datasets.

\subsection{Effects of Training Data Size}
\label{sec:trsize}

\begin{figure}[t]
	\centering
	\includegraphics[width=0.47\textwidth]{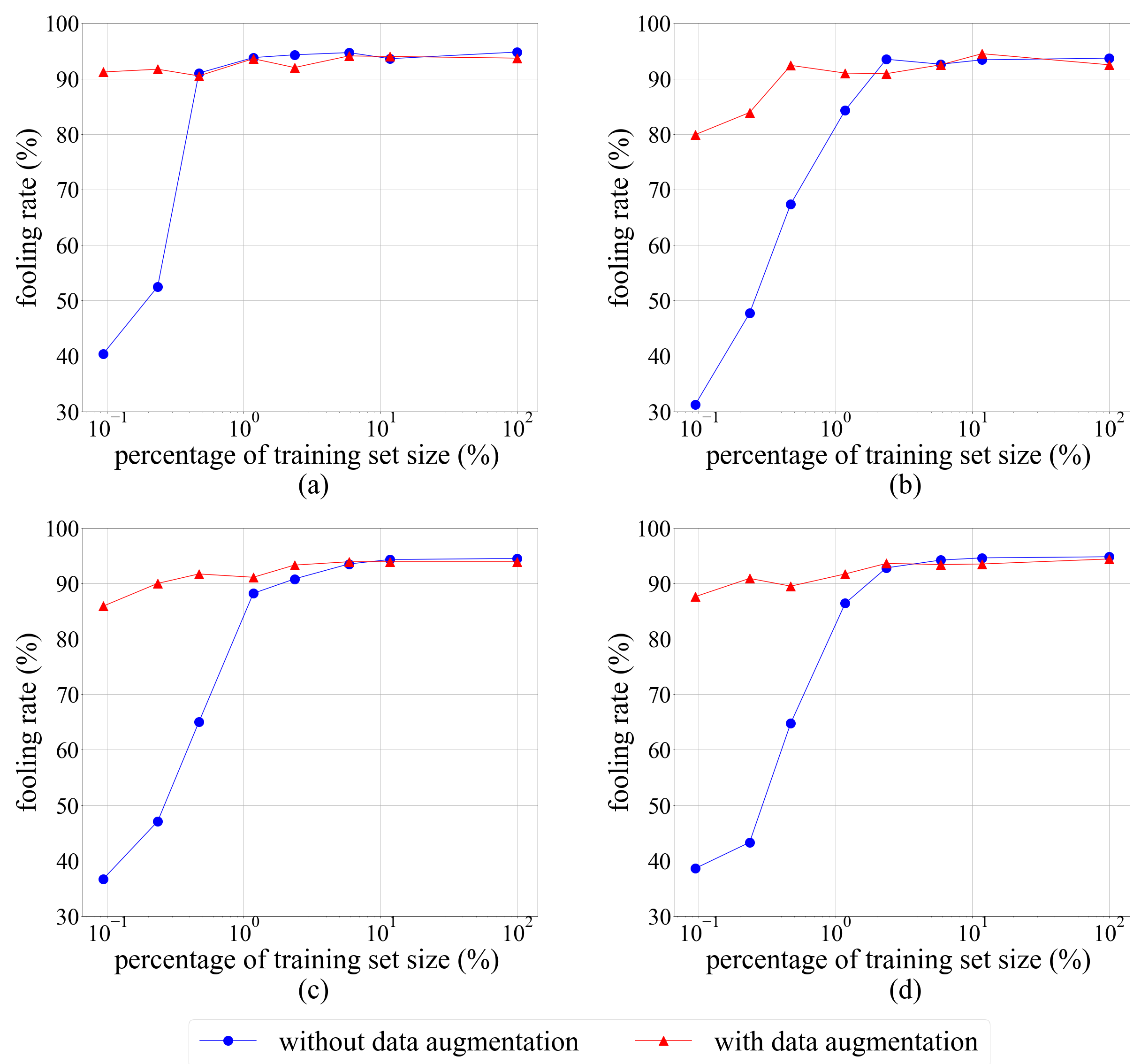}
	\caption{Fooling rate in function of training set size (expressed as a percentage of the full training set size) using the MINC dataset and various DNN models ((a) ResNet~\cite{he2016deep}, (b) DeepTEN~\cite{zhang2017deep}, (c) DEP~\cite{xue2018deep} and (d) MuLTER~\cite{hu2019multi}) where the training set is the set used to compute the universal perturbations.}
	\label{fig:trsize}
\end{figure}

We explore the effects of the training data size on the universal perturbation effectiveness for our FTGAP method using the MINC dataset, which includes 48,875 training images. According to Figure~\ref{fig:trsize}, without data augmentation, the fooling rates can be maintained at the same level when the percentage of training images that are used to compute the universal perturbations drops from 100\% to 1\%. The achieved fooling rate drastically decreases when the percentage of training set size goes lower than 1\%. When data augmentation is employed\footnote{We employed the same data augmentation approaches as the ones that were used to train the original DNN texture classifiers~\cite{krizhevsky2012imagenet, zhang2017deep}.}, the proposed FTGAP can maintain a relatively high fooling rate with only a marginal decrease (about 10\% drop at most) even when the universal perturbation is computed with about 0.1\% of all training images, indicating a strong robustness against very sparse data. Similar results were also obtained for other datasets (see Appendix~\ref{sec:trsize_a}).

\section{Conclusion}

To our knowledge this paper presents the first work dealing with universal adversarial attacks on texture recognition tasks. To this end, this paper presents a novel frequency-tuned generative adversarial perturbation (FTGAP) method. The proposed FTGAP method generates universal adversarial perturbations by computing frequency-domain perturbation components that are adapted to the local characteristics of DCT frequency bands. Furthermore, we demonstrate that, compared to existing universal attack methods, our method can significantly reduce the perceptibility of the generated universal perturbations while achieving on average comparable or higher fooling rates across datasets and models for texture image classification. In addition, according to the conducted experiments, the proposed FTGAP method can improve the universal attack performance as compared to existing universal attacks in terms of various aspects including out-of-distribution generalization across datasets and robustness against defended models, and can also maintain a relatively high fooling rate even with a very sparse set of training samples.

{\small
	\bibliographystyle{ieee_fullname}
	\bibliography{egbib}

\begin{thebibliography}{10}\itemsep=-1pt

\bibitem{ahumada1992luminance}
Albert~J Ahumada~Jr and Heidi~A Peterson.
\newblock Luminance-model-based {DCT} quantization for color image compression.
\newblock In {\em Human Vision, Visual Processing, and Digital Display III},
  volume 1666, pages 365--374, 1992.

\bibitem{ANDREARCZYK201663}
Vincent Andrearczyk and Paul~F. Whelan.
\newblock Using filter banks in convolutional neural networks for texture
  classification.
\newblock {\em Pattern Recognition Letters}, 84:63 -- 69, 2016.

\bibitem{bell2015material}
Sean Bell, Paul Upchurch, Noah Snavely, and Kavita Bala.
\newblock Material recognition in the wild with the materials in context
  database.
\newblock In {\em IEEE conference on Computer Vision and Pattern Recognition},
  pages 3479--3487, 2015.

\bibitem{caputo2005class}
Barbara Caputo, Eric Hayman, and P Mallikarjuna.
\newblock Class-specific material categorisation.
\newblock In {\em IEEE International Conference on Computer Vision}, volume~2,
  pages 1597--1604, 2005.

\bibitem{carlini2017towards}
Nicholas Carlini and David Wagner.
\newblock Towards evaluating the robustness of neural networks.
\newblock In {\em IEEE Symposium on Security and Privacy}, pages 39--57, 2017.

\bibitem{cimpoi2014describing}
Mircea Cimpoi, Subhransu Maji, Iasonas Kokkinos, Sammy Mohamed, and Andrea
  Vedaldi.
\newblock Describing textures in the wild.
\newblock In {\em IEEE Conference on Computer Vision and Pattern Recognition},
  pages 3606--3613, 2014.

\bibitem{cimpoi2015deep}
Mircea Cimpoi, Subhransu Maji, and Andrea Vedaldi.
\newblock Deep filter banks for texture recognition and segmentation.
\newblock In {\em IEEE Conference on Computer Vision and Pattern Recognition},
  pages 3828--3836, 2015.

\bibitem{croce2019sparse}
Francesco Croce and Matthias Hein.
\newblock Sparse and imperceivable adversarial attacks.
\newblock In {\em IEEE International Conference on Computer Vision}, pages
  4724--4732, 2019.

\bibitem{csurka2004visual}
Gabriella Csurka, Christopher Dance, Lixin Fan, Jutta Willamowski, and
  C{\'e}dric Bray.
\newblock Visual categorization with bags of keypoints.
\newblock In {\em Workshop on statistical learning in computer vision, ECCV},
  volume~1, pages 1--2. Prague, 2004.

\bibitem{dabouei2020smoothfool}
Ali Dabouei, Sobhan Soleymani, Fariborz Taherkhani, Jeremy Dawson, and Nasser
  Nasrabadi.
\newblock {SmoothFool}: an efficient framework for computing smooth adversarial
  perturbations.
\newblock In {\em IEEE Winter Conference on Applications of Computer Vision},
  pages 2665--2674, 2020.

\bibitem{dai2017fason}
Xiyang Dai, Joe Yue-Hei~Ng, and Larry~S Davis.
\newblock {FASON}: first and second order information fusion network for
  texture recognition.
\newblock In {\em IEEE Conference on Computer Vision and Pattern Recognition},
  pages 7352--7360, 2017.

\bibitem{deng2020frequency}
Yingpeng Deng and Lina~J Karam.
\newblock Frequency-tuned universal adversarial attacks.
\newblock {\em arXiv preprint arXiv:2003.05549}, 2020.

\bibitem{9191288}
Yingpeng Deng and Lina~J. Karam.
\newblock Universal adversarial attack via enhanced projected gradient descent.
\newblock In {\em IEEE International Conference on Image Processing}, pages
  1241--1245, 2020.

\bibitem{dong2018boosting}
Yinpeng Dong, Fangzhou Liao, Tianyu Pang, Hang Su, Jun Zhu, Xiaolin Hu, and
  Jianguo Li.
\newblock Boosting adversarial attacks with momentum.
\newblock In {\em IEEE Conference on Computer Vision and Pattern Recognition},
  pages 9185--9193, 2018.

\bibitem{dong2019evading}
Yinpeng Dong, Tianyu Pang, Hang Su, and Jun Zhu.
\newblock Evading defenses to transferable adversarial examples by
  translation-invariant attacks.
\newblock In {\em IEEE Conference on Computer Vision and Pattern Recognition},
  pages 4312--4321, 2019.

\bibitem{fan2020sparse}
Yanbo Fan, Baoyuan Wu, Tuanhui Li, Yong Zhang, Mingyang Li, Zhifeng Li, and
  Yujiu Yang.
\newblock Sparse adversarial attack via perturbation factorization.
\newblock In {\em European Conference on Computer Vision}, 2020.

\bibitem{freeman1997learning}
William~T Freeman and Joshua~B Tenenbaum.
\newblock Learning bilinear models for two-factor problems in vision.
\newblock In {\em IEEE Conference on Computer Vision and Pattern Recognition},
  pages 554--560. IEEE, 1997.

\bibitem{girshick2014rich}
Ross Girshick, Jeff Donahue, Trevor Darrell, and Jitendra Malik.
\newblock Rich feature hierarchies for accurate object detection and semantic
  segmentation.
\newblock In {\em IEEE Conference on Computer Vision and Pattern Recognition},
  pages 580--587, 2014.

\bibitem{goodfellow2014generative}
Ian Goodfellow, Jean Pouget-Abadie, Mehdi Mirza, Bing Xu, David Warde-Farley,
  Sherjil Ozair, Aaron Courville, and Yoshua Bengio.
\newblock Generative adversarial nets.
\newblock In {\em Advances in Neural Information Processing Systems}, pages
  2672--2680, 2014.

\bibitem{goodfellow2014explaining}
Ian~J Goodfellow, Jonathon Shlens, and Christian Szegedy.
\newblock Explaining and harnessing adversarial examples.
\newblock {\em International Conference on Learning Representations}, 2015.

\bibitem{guo2018low}
Chuan Guo, Jared~S Frank, and Kilian~Q Weinberger.
\newblock Low frequency adversarial perturbation.
\newblock {\em arXiv preprint arXiv:1809.08758}, 2018.

\bibitem{guo2019simple}
Chuan Guo, Jacob Gardner, Yurong You, Andrew~Gordon Wilson, and Kilian
  Weinberger.
\newblock Simple black-box adversarial attacks.
\newblock In {\em International Conference on Machine Learning}, pages
  2484--2493, 2019.

\bibitem{he2016deep}
Kaiming He, Xiangyu Zhang, Shaoqing Ren, and Jian Sun.
\newblock Deep residual learning for image recognition.
\newblock In {\em IEEE Conference on Computer Vision and Pattern Recognition},
  pages 770--778, 2016.

\bibitem{hendrik2017universal}
Jan Hendrik~Metzen, Mummadi Chaithanya~Kumar, Thomas Brox, and Volker Fischer.
\newblock Universal adversarial perturbations against semantic image
  segmentation.
\newblock In {\em IEEE International Conference on Computer Vision}, pages
  2755--2764, 2017.

\bibitem{hontsch2002adaptive}
Ingo Hontsch and Lina~J Karam.
\newblock Adaptive image coding with perceptual distortion control.
\newblock {\em IEEE Transactions on Image Processing}, 11(3):213--222, 2002.

\bibitem{hu2019multi}
Yuting Hu, Zhiling Long, and Ghassan AlRegib.
\newblock Multi-level texture encoding and representation ({M}u{LTER}) based on
  deep neural networks.
\newblock In {\em IEEE International Conference on Image Processing}, pages
  4410--4414, 2019.

\bibitem{jegou2010aggregating}
Herv{\'e} J{\'e}gou, Matthijs Douze, Cordelia Schmid, and Patrick P{\'e}rez.
\newblock Aggregating local descriptors into a compact image representation.
\newblock In {\em IEEE Conference on Computer Vision and Pattern Recognition},
  pages 3304--3311, 2010.

\bibitem{karam2011efficient}
Lina~J Karam, Nabil~G Sadaka, Rony Ferzli, and Zoran~A Ivanovski.
\newblock An efficient selective perceptual-based super-resolution estimator.
\newblock {\em IEEE Transactions on Image Processing}, 20(12):3470--3482, 2011.

\bibitem{kingma2014adam}
Diederik~P Kingma and Jimmy Ba.
\newblock Adam: A method for stochastic optimization.
\newblock {\em arXiv preprint arXiv:1412.6980}, 2014.

\bibitem{krizhevsky2012imagenet}
Alex Krizhevsky, Ilya Sutskever, and Geoffrey~E Hinton.
\newblock Imagenet classification with deep convolutional neural networks.
\newblock In {\em Advances in Neural Information Processing Systems}, pages
  1097--1105, 2012.

\bibitem{kurakin2016adversarial}
Alexey Kurakin, Ian Goodfellow, and Samy Bengio.
\newblock Adversarial examples in the physical world.
\newblock {\em arXiv preprint arXiv:1607.02533}, 2016.

\bibitem{li2020universal}
Debang Li, Junge Zhang, and Kaiqi Huang.
\newblock Universal adversarial perturbations against object detection.
\newblock {\em Pattern Recognition}, page 107584, 2020.

\bibitem{li2019regional}
Yingwei Li, Song Bai, Cihang Xie, Zhenyu Liao, Xiaohui Shen, and Alan Yuille.
\newblock Regional homogeneity: Towards learning transferable universal
  adversarial perturbations against defenses.
\newblock {\em European Conference on Computer Vision}, 2020.

\bibitem{lin2016visualizing}
Tsung-Yu Lin and Subhransu Maji.
\newblock Visualizing and understanding deep texture representations.
\newblock In {\em IEEE Conference on Computer Vision and Pattern Recognition},
  pages 2791--2799, 2016.

\bibitem{lin2015bilinear}
Tsung-Yu Lin, Aruni RoyChowdhury, and Subhransu Maji.
\newblock Bilinear {CNN} models for fine-grained visual recognition.
\newblock In {\em IEEE International Conference on Computer Vision}, pages
  1449--1457, 2015.

\bibitem{liu2019universal}
Hong Liu, Rongrong Ji, Jie Li, Baochang Zhang, Yue Gao, Yongjian Wu, and Feiyue
  Huang.
\newblock Universal adversarial perturbation via prior driven uncertainty
  approximation.
\newblock In {\em IEEE International Conference on Computer Vision}, pages
  2941--2949, 2019.

\bibitem{liu2019geometry}
Yujia Liu, Seyed-Mohsen Moosavi-Dezfooli, and Pascal Frossard.
\newblock A geometry-inspired decision-based attack.
\newblock In {\em IEEE International Conference on Computer Vision}, pages
  4890--4898, 2019.

\bibitem{liu2006jpeg2000}
Zhen Liu, Lina~J Karam, and Andrew~B Watson.
\newblock {JPEG2000} encoding with perceptual distortion control.
\newblock {\em IEEE Transactions on Image Processing}, 15(7):1763--1778, 2006.

\bibitem{long2015fully}
Jonathan Long, Evan Shelhamer, and Trevor Darrell.
\newblock Fully convolutional networks for semantic segmentation.
\newblock In {\em IEEE Conference on Computer Vision and Pattern Recognition},
  pages 3431--3440, 2015.

\bibitem{lowe2004distinctive}
David~G Lowe.
\newblock Distinctive image features from scale-invariant keypoints.
\newblock {\em International Journal of Computer Vision}, 60(2):91--110, 2004.

\bibitem{luo2018towards}
Bo Luo, Yannan Liu, Lingxiao Wei, and Qiang Xu.
\newblock Towards imperceptible and robust adversarial example attacks against
  neural networks.
\newblock In {\em AAAI Conference on Artificial Intelligence}, 2018.

\bibitem{modas2019sparsefool}
Apostolos Modas, Seyed-Mohsen Moosavi-Dezfooli, and Pascal Frossard.
\newblock {SparseFool}: a few pixels make a big difference.
\newblock In {\em IEEE Conference on Computer Vision and Pattern Recognition},
  pages 9087--9096, 2019.

\bibitem{moosavi2017universal}
Seyed-Mohsen Moosavi-Dezfooli, Alhussein Fawzi, Omar Fawzi, and Pascal
  Frossard.
\newblock Universal adversarial perturbations.
\newblock In {\em IEEE Conference on Computer Vision and Pattern Recognition},
  pages 1765--1773, 2017.

\bibitem{moosavi2016deepfool}
Seyed-Mohsen Moosavi-Dezfooli, Alhussein Fawzi, and Pascal Frossard.
\newblock {DeepFool: A} simple and accurate method to fool deep neural
  networks.
\newblock In {\em IEEE Conference on Computer Vision and Pattern Recognition},
  pages 2574--2582, 2016.

\bibitem{mopuri2018generalizable}
Konda~Reddy Mopuri, Aditya Ganeshan, and R~Venkatesh Babu.
\newblock Generalizable data-free objective for crafting universal adversarial
  perturbations.
\newblock {\em IEEE Transactions on Pattern Analysis and Machine Intelligence},
  41(10):2452--2465, 2018.

\bibitem{mopuri-bmvc-2017}
Konda~Reddy Mopuri, Utsav Garg, and R~Venkatesh Babu.
\newblock {Fast Feature Fool}: a data independent approach to universal
  adversarial perturbations.
\newblock In {\em British Machine Vision Conference}, 2017.

\bibitem{mummadi2019defending}
Chaithanya~Kumar Mummadi, Thomas Brox, and Jan~Hendrik Metzen.
\newblock Defending against universal perturbations with shared adversarial
  training.
\newblock In {\em IEEE International Conference on Computer Vision}, pages
  4928--4937, 2019.

\bibitem{papernot2016limitations}
Nicolas Papernot, Patrick McDaniel, Somesh Jha, Matt Fredrikson, Z~Berkay
  Celik, and Ananthram Swami.
\newblock The limitations of deep learning in adversarial settings.
\newblock In {\em IEEE European Symposium on Security and Privacy}, pages
  372--387, 2016.

\bibitem{perronnin2010improving}
Florent Perronnin, Jorge S{\'a}nchez, and Thomas Mensink.
\newblock Improving the fisher kernel for large-scale image classification.
\newblock In {\em European Conference on Computer Vision}, pages 143--156.
  Springer, 2010.

\bibitem{poursaeed2018generative}
Omid Poursaeed, Isay Katsman, Bicheng Gao, and Serge Belongie.
\newblock Generative adversarial perturbations.
\newblock In {\em IEEE Conference on Computer Vision and Pattern Recognition},
  pages 4422--4431, 2018.

\bibitem{reddy2018ask}
Konda Reddy~Mopuri, Phani Krishna~Uppala, and R Venkatesh~Babu.
\newblock Ask, acquire, and attack: Data-free {UAP} generation using class
  impressions.
\newblock In {\em European Conference on Computer Vision}, pages 19--34, 2018.

\bibitem{reddy2018nag}
Konda Reddy~Mopuri, Utkarsh Ojha, Utsav Garg, and R Venkatesh~Babu.
\newblock {NAG}: network for adversary generation.
\newblock In {\em IEEE Conference on Computer Vision and Pattern Recognition},
  pages 742--751, 2018.

\bibitem{redmon2016you}
Joseph Redmon, Santosh Divvala, Ross Girshick, and Ali Farhadi.
\newblock You only look once: Unified, real-time object detection.
\newblock In {\em IEEE Conference on Computer Vision and Pattern Recognition},
  pages 779--788, 2016.

\bibitem{ronneberger2015u}
Olaf Ronneberger, Philipp Fischer, and Thomas Brox.
\newblock {U-Net: Convolutional} networks for biomedical image segmentation.
\newblock In {\em International Conference on Medical Image Computing and
  Computer-Assisted Intervention}, pages 234--241, 2015.

\bibitem{ILSVRC15}
Olga Russakovsky, Jia Deng, Hao Su, Jonathan Krause, Sanjeev Satheesh, Sean Ma,
  Zhiheng Huang, Andrej Karpathy, Aditya Khosla, Michael Bernstein,
  Alexander~C. Berg, and Li Fei-Fei.
\newblock {ImageNet} large scale visual recognition challenge.
\newblock {\em International Journal of Computer Vision}, 115(3):211--252,
  2015.

\bibitem{shafahi2020universal}
Ali Shafahi, Mahyar Najibi, Zheng Xu, John Dickerson, Larry~S Davis, and Tom
  Goldstein.
\newblock Universal adversarial training.
\newblock {\em AAAI Conference on Artificial Intelligence}, 2020.

\bibitem{shamsabadi2020colorfool}
Ali~Shahin Shamsabadi, Ricardo Sanchez-Matilla, and Andrea Cavallaro.
\newblock Colorfool: Semantic adversarial colorization.
\newblock In {\em Proceedings of the IEEE/CVF Conference on Computer Vision and
  Pattern Recognition}, pages 1151--1160, 2020.

\bibitem{sharan2013recognizing}
Lavanya Sharan, Ce Liu, Ruth Rosenholtz, and Edward~H Adelson.
\newblock Recognizing materials using perceptually inspired features.
\newblock {\em International Journal of Computer Vision}, 103(3):348--371,
  2013.

\bibitem{sharif2018suitability}
Mahmood Sharif, Lujo Bauer, and Michael~K Reiter.
\newblock On the suitability of lp-norms for creating and preventing
  adversarial examples.
\newblock In {\em IEEE Conference on Computer Vision and Pattern Recognition
  Workshops}, pages 1605--1613, 2018.

\bibitem{sharma2019effectiveness}
Yash Sharma, Gavin~Weiguang Ding, and Marcus~A Brubaker.
\newblock On the effectiveness of low frequency perturbations.
\newblock In {\em International Joint Conference on Artificial Intelligence},
  pages 3389--3396, 2019.

\bibitem{Simonyan15}
Karen Simonyan and Andrew Zisserman.
\newblock Very deep convolutional networks for large-scale image recognition.
\newblock In {\em International Conference on Learning Representations}, 2015.

\bibitem{su2019one}
Jiawei Su, Danilo~Vasconcellos Vargas, and Kouichi Sakurai.
\newblock One pixel attack for fooling deep neural networks.
\newblock {\em IEEE Transactions on Evolutionary Computation}, 23(5):828--841,
  2019.

\bibitem{szegedy2015going}
Christian Szegedy, Wei Liu, Yangqing Jia, Pierre Sermanet, Scott Reed, Dragomir
  Anguelov, Dumitru Erhan, Vincent Vanhoucke, and Andrew Rabinovich.
\newblock Going deeper with convolutions.
\newblock In {\em IEEE Conference on Computer Vision and Pattern Recognition},
  pages 1--9, 2015.

\bibitem{szegedy2013intriguing}
Christian Szegedy, Wojciech Zaremba, Ilya Sutskever, Joan Bruna, Dumitru Erhan,
  Ian Goodfellow, and Rob Fergus.
\newblock Intriguing properties of neural networks.
\newblock {\em International Conference on Learning Representations}, 2014.

\bibitem{tsuzuku2019structural}
Yusuke Tsuzuku and Issei Sato.
\newblock On the structural sensitivity of deep convolutional networks to the
  directions of fourier basis functions.
\newblock In {\em IEEE Conference on Computer Vision and Pattern Recognition},
  pages 51--60, 2019.

\bibitem{wang20164d}
Ting-Chun Wang, Jun-Yan Zhu, Ebi Hiroaki, Manmohan Chandraker, Alexei~A Efros,
  and Ravi Ramamoorthi.
\newblock A {4D} light-field dataset and {CNN} architectures for material
  recognition.
\newblock In {\em European Conference on Computer Vision}, pages 121--138,
  2016.

\bibitem{wang2004image}
Zhou Wang, Alan~C Bovik, Hamid~R Sheikh, and Eero~P Simoncelli.
\newblock Image quality assessment: from error visibility to structural
  similarity.
\newblock {\em IEEE Transactions on Image Processing}, 13(4):600--612, 2004.

\bibitem{wang2019invisible}
Zhibo Wang, Mengkai Song, Siyan Zheng, Zhifei Zhang, Yang Song, and Qian Wang.
\newblock Invisible adversarial attack against deep neural networks: An
  adaptive penalization approach.
\newblock {\em IEEE Transactions on Dependable and Secure Computing}, 2019.

\bibitem{xie2017adversarial}
Cihang Xie, Jianyu Wang, Zhishuai Zhang, Yuyin Zhou, Lingxi Xie, and Alan
  Yuille.
\newblock Adversarial examples for semantic segmentation and object detection.
\newblock In {\em IEEE International Conference on Computer Vision}, pages
  1369--1378, 2017.

\bibitem{xie2019improving}
Cihang Xie, Zhishuai Zhang, Yuyin Zhou, Song Bai, Jianyu Wang, Zhou Ren, and
  Alan~L Yuille.
\newblock Improving transferability of adversarial examples with input
  diversity.
\newblock In {\em IEEE Conference on Computer Vision and Pattern Recognition},
  pages 2730--2739, 2019.

\bibitem{xu2018structured}
Kaidi Xu, Sijia Liu, Pu Zhao, Pin-Yu Chen, Huan Zhang, Quanfu Fan, Deniz
  Erdogmus, Yanzhi Wang, and Xue Lin.
\newblock Structured adversarial attack: Towards general implementation and
  better interpretability.
\newblock In {\em International Conference on Learning Representations}, 2019.

\bibitem{xue2018deep}
Jia Xue, Hang Zhang, and Kristin Dana.
\newblock Deep texture manifold for ground terrain recognition.
\newblock In {\em IEEE Conference on Computer Vision and Pattern Recognition},
  pages 558--567, 2018.

\bibitem{xue2017differential}
Jia Xue, Hang Zhang, Kristin Dana, and Ko Nishino.
\newblock Differential angular imaging for material recognition.
\newblock In {\em IEEE Conference on Computer Vision and Pattern Recognition},
  pages 764--773, 2017.

\bibitem{yin2019fourier}
Dong Yin, Raphael~Gontijo Lopes, Jon Shlens, Ekin~Dogus Cubuk, and Justin
  Gilmer.
\newblock A {Fourier} perspective on model robustness in computer vision.
\newblock In {\em Advances in Neural Information Processing Systems}, pages
  13276--13286, 2019.

\bibitem{zhang2017deep}
Hang Zhang, Jia Xue, and Kristin Dana.
\newblock Deep {TEN}: Texture encoding network.
\newblock In {\em IEEE Conference on Computer Vision and Pattern Recognition},
  pages 708--717, 2017.

\bibitem{zhang2020advjnd}
Zifei Zhang, Kai Qiao, Lingyun Jiang, Linyuan Wang, and Bin Yan.
\newblock Adv{JND}: Generating adversarial examples with just noticeable
  difference.
\newblock {\em arXiv preprint arXiv:2002.00179}, 2020.

\bibitem{zhao2020towards}
Zhengyu Zhao, Zhuoran Liu, and Martha Larson.
\newblock Towards large yet imperceptible adversarial image perturbations with
  perceptual color distance.
\newblock In {\em IEEE Conference on Computer Vision and Pattern Recognition},
  pages 1039--1048, 2020.

\bibitem{zhou2018transferable}
Wen Zhou, Xin Hou, Yongjun Chen, Mengyun Tang, Xiangqi Huang, Xiang Gan, and
  Yong Yang.
\newblock Transferable adversarial perturbations.
\newblock In {\em European Conference on Computer Vision}, pages 452--467,
  2018.

\bibitem{zujovic2013structural}
Jana Zujovic, Thrasyvoulos~N Pappas, and David~L Neuhoff.
\newblock Structural texture similarity metrics for image analysis and
  retrieval.
\newblock {\em IEEE Transactions on Image Processing}, 22(7):2545--2558, 2013.

\end{thebibliography}
}

\newpage

\appendix

\renewcommand{\appendixpagename}{\twocolumn[\centering Appendix\\ \ \\]}

\appendixpage

\section{Computation Details of JND Thresholds}
\label{sec:compjnd}

The JND thresholds for different frequency bands $(k_1, k_2)$ can be computed as~\cite{hontsch2002adaptive}
\begin{equation}
\small
t_{DCT}(k_1,k_2) = \frac{MT(k_1,k_2)}{2\tilde{c_1}(k_1)\tilde{c_2}(k_2)(L_{max}-L_{min})},
\label{eqtdct}
\end{equation}
where $L_{min}$ and $L_{max}$ are the minimum and maximum display luminance, $M=255$ for 8-bit image, and $\tilde{c_i(k_i)}$ are computed using Equation~\ref{eqdct3} in the main paper. To compute the background luminance-adjusted contrast sensitivity $T(k_1,k_2)$, Ahumada and Peterson proposed an approximating parametric model\footnote[1]{$T(k_1,k_2)$ can be computed for any $k_1$, $k_2$ which satisfy $k_1k_2\neq0$ by this model, while $T(0,0)$ is estimated as min($T(0,1)$, $T(1,0)$).}~\cite{ahumada1992luminance}:
\begin{equation}
\small
\begin{aligned}
\log_{10}(T(k_1,k_2))=&\log_{10}\frac{T_{min}}{r+(1-r)\cos^2\theta(k_1,k_2)}\\
&+K(\log_{10}f(k_1,k_2)-\log_{10}f_{min})^2.
\end{aligned}
\label{eqlogt}
\end{equation}
The radial frequency $f(k_1,k_2)$ and its corresponding orientation $\theta(k_1,k_2)$ are given as follows:
\begin{equation}
\small
f(k_1,k_2) = \frac{1}{2N_{DCT}}\sqrt{\frac{k_1^2}{w_x^2}+\frac{k_2^2}{w_y^2}},
\label{eqf}
\end{equation}
\begin{equation}
\small
\theta(k_1,k_2)=\arcsin\frac{2f(k_1,0)f(0,k_2)}{f^2(k_1,k_2)}.
\label{eqtheta}
\end{equation}
The luminance-dependent parameters are generated by the following equations:
\begin{equation}
\small
T_{min}=\left\{
\begin{array}{cc}
\left(\frac{L}{L_T}\right)^{\alpha_T}\frac{L_T}{S_0}&,\ L\leq L_T  \\
\frac{L}{S_0}&,\ L>L_T
\end{array}
\right., \ 
\label{eqtmin}
\end{equation}
\begin{equation}
\small
f_{min}=\left\{
\begin{array}{cc}
f_0\left(\frac{L}{L_f}\right)^{\alpha_f}&,\ L\leq L_f  \\
f_0&,\ L>L_f
\end{array}
\right.,\
\label{eqfmin}
\end{equation}
\begin{equation}
K=\left\{
\begin{array}{cc}
K_0\left(\frac{L}{L_K}\right)^{\alpha_K}&,\ L\leq L_K  \\
K_0&,\ L>L_K
\end{array}
\right..
\label{eqk}
\end{equation}
The values of constants in Equations~\ref{eqtmin}-\ref{eqk} are $r=0.7$, $N_{DCT}=8$, $L_T=13.45$ cd/m$^2$, $S_0=94.7$, $\alpha_T=0.649$, $f_0=6.78$ cycles/degree, $alpha_f=0.182$, $L_f=300$ cd/m$^2$, $K_0=3.125$, $\alpha_K=0.0706$, and $L_K=300$ cd/m$^2$. Given a viewing distance of 60 cm and a 31.5 pixels-per-cm (80 pixels-per-inch) display, the horizontal width/vertical height of a pixel ($w_x/w_y$) is 0.0303 degree of visual angle~\cite{liu2006jpeg2000}. In practice, for a measured luminance of $L_{min}=0$ cd/m$^2$ and $L_{max}=175$ cd/m$^2$, we use the luminance $L$ corresponding to the median intensity value of the image to avoid image-specific computation as follows~\cite{karam2011efficient}:
\begin{equation}
L=L_{min}+128\frac{L_{max}-L_{min}}{M}
\label{eql}
\end{equation}

\section{Implementation Details}
\label{sec:imple}

In this section, we will describe all the adopted texture datasets and corresponding splits for training/attacking/testing. The details for training the DNN texture classifiers are also provided together with the hyperparameter settings for our proposed FTGAP method. 

\subsection{Datasets}

\begin{table}
	\centering
	\caption{The number of images used for training the networks (training set), computing the perturbation (attacking set) and testing (testing set).}
	\label{imgnums}
	\resizebox{0.4\textwidth}{!}{
		\begin{tabular}[t]{|c|c|c|c|}
			\hline
			Dataset&training set&attacking set&testing set\\
			\hline
			MINC~\cite{bell2015material}&48,875&5,750&5,750\\
			\hline
			GTOS~\cite{xue2017differential, xue2018deep}&93,945&9,381&6,066\\
			\hline
			DTD~\cite{cimpoi2014describing}&3,760&1,880&1,880\\
			\hline
			4DLF~\cite{wang20164d}&840&360&360\\
			\hline
			FMD~\cite{sharan2013recognizing}&900&300&100\\
			\hline
			KTH~\cite{caputo2005class}&2,376&231&2,376\\
			\hline
	\end{tabular}}
\end{table}

We consider six texture datasets in order to examine the recognition performance of the DNN models under various unviersal attacks. The Materials in Context (MINC) Database~\cite{bell2015material} is a large real-world material dataset. In our work, we adopt its publicly available subset MINC-2500 with its provided train-test split. There are 23 classes with 2500 images for each class. Xue \textit{et al.} created the Ground Terrain in Outdoor Scenes (GTOS)~\cite{xue2017differential} dataset with 31 classes of over 90,000 ground terrain images and the GTOS-mobile ~\cite{xue2018deep} with the same classes but with a much smaller number of images (around 6,000 images). As in~\cite{xue2018deep}, we adopt the GTOS dataset as the training set and test the trained models on the GTOS-mobile dataset. In our paper, we use MINC and GTOS to refer to the MINC-2500 and the combined dataset of GTOS and GTOS-mobile, respectively.

We also adopt other smaller texture datasets. The Describable Textures Database (DTD)~\cite{cimpoi2014describing} includes 47 categories with 120 images per category. The 4D light-field (4DLF) material dataset~\cite{wang20164d} consists of 1200 images in total for 12 different categories. An angular resolution of $7 \times 7$ is used for each image in the dataset and we only use the one where $(u, v) = (-3, 3)$ in our experiments. The Filckr Material Dataset (FMD)~\cite{sharan2013recognizing} has 10 material classes and 100 images per class. The KTH-TIPS-2b (KTH)~\cite{caputo2005class} comprises 11 texture classes, with four samples per class and 108 images per sample.

For each dataset, the training set is used to train the models for texture recognition, the attacking set includes images that are randomly sampled from the training set for computing the perturbations, and the testing set is to evalute the performance. The number of images for each of these sets is listed in Table~\ref{imgnums}. We use the train-test split that is either provided in the dataset or suggested in~\cite{zhang2017deep}. To extract the attacking set, we randomly sample 10\% of the images in the training image set for the GTOS/KTH dataset, one-third of the training images for the FMD dataset and a number of training images that is equal to the number of testing images for MINC, DTD and 4DLF separately.

\subsection{Training Strategies}

\begin{table}
	\centering
	\caption{The hyperparameter settings on different datasets for FTGAP.}
	\label{hp}
	\resizebox{0.4\textwidth}{!}{
		\begin{tabular}[t]{|c|cccccc|}
			\hline
			&MINC&GTOS&DTD$^*$&4DLF$^*$&FMD&KTH$^*$\\
			\hline
			$\lambda_l$&0&1&0&0&0&0\\
			\hline
			$\lambda_h$&3&3&1.5&2.5&2.5&2\\
			\hline
			$f_c$&\multicolumn{6}{c|}{4}\\
			\hline
	\end{tabular}}
\end{table}

\begin{table}[t]
	\centering
	\caption{Top-1 accuracy results for attacking the defended models on MINC and GTOS using GAP-tar and our FTGAP.}
	\label{robustuat}
	
	\resizebox{0.47\textwidth}{!}{
		\begin{tabular}{|c|c|c|cccc|c|}
			\hline
			\multicolumn{2}{|c|}{ }&no attack&ResNet&DeepTEN&DEP&MuLTER&mean\\
			\hline
			\multirow{2}*{MINC}&GAP-tar&\multirow{2}*{78.6}&27.6&17.1&11.8&14.4&17.7\\
			~&FTGAP&~&\textbf{5.0}&\textbf{5.9}&\textbf{5.9}&\textbf{5.5}&\textbf{5.4}\\
			\hline
			\multirow{2}*{GTOS}&GAP-tar&\multirow{2}*{73.3}&48.8&56.2&54.1&48.0&51.8\\
			~&FTGAP&~&\textbf{48.6}&\textbf{28.8}&\textbf{42.4}&\textbf{38.1}&\textbf{39.5}\\
			\hline
	\end{tabular}}
\end{table}

The DNN models are finetuned based on the ResNet backbones which are pretrained on ImageNet~\cite{ILSVRC15}. With regard to the network backbone, we use pretrained ResNet50\footnote[2]{According to the provided source code for~\cite{zhang2017deep}, the adopted ResNet50 backbones are slightly different from~\cite{he2016deep}, where the first convolutional layer with a kernel size of 7 is replaced by three cascaded 3 $\times$ 3 convolutional layers.} for MINC, DTD, 4DLF and FMD and pretrained ResNet18 for GTOS and KTH, as suggested in~\cite{zhang2017deep, xue2018deep}. Following the same training strategies in~\cite{xue2018deep, hu2019multi}, we train our models with only single-size images. Similar to the data augmentation strategies in~\cite{zhang2017deep}, the input images are resized to 256 $\times$ 256 and then randomly cropped to 224 $\times$ 224, followed by a random horizontal flipping. Standard color augmentation and PCA-based noise are used as in~\cite{krizhevsky2012imagenet, zhang2017deep}. For DeepTEN~\cite{zhang2017deep}, DEP~\cite{xue2018deep} and MuLTER~\cite{hu2019multi}, the number of codewords is set to 8 for the ResNet18 backbone and to 32 for the ResNet50 backbone as suggested by the corresponding authors. For finetuning, we use stochastic gradient descent (SGD) with a mini-batch size of 32. With a weight decay of $10^{-4}$ and a momentum of 0.9, the learning rate is initialized to 0.01 and decays every 10 epochs by a factor of 0.1. The training process is terminated after 30 epochs and the best performing model is adopted.

\begin{figure}[tb]
	\centering
	
	\begin{minipage}{0.11\textwidth}
		\centering
		\small
		ResNet~\cite{he2016deep}
		\includegraphics[width=1\textwidth]{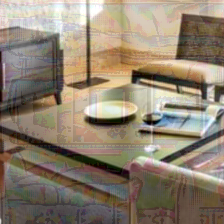}\\	
		\includegraphics[width=1\textwidth]{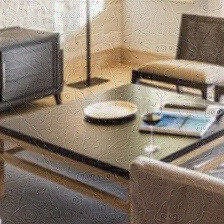}\\
		\ \\
		\includegraphics[width=1\textwidth]{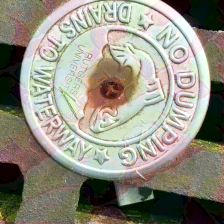}\\	
		\includegraphics[width=1\textwidth]{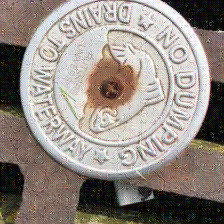}\\
	\end{minipage}
	\begin{minipage}{0.11\textwidth}
		\centering
		\small
		DeepTEN~\cite{zhang2017deep}
		\includegraphics[width=1\textwidth]{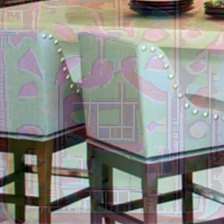}\\	
		\includegraphics[width=1\textwidth]{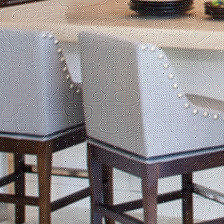}\\
		\ \\
		\includegraphics[width=1\textwidth]{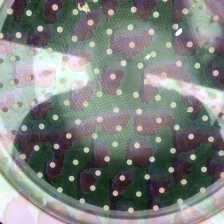}\\	
		\includegraphics[width=1\textwidth]{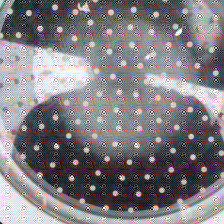}\\
	\end{minipage}
	\begin{minipage}{0.11\textwidth}
		\centering
		\small
		DEP~\cite{xue2018deep}
		\includegraphics[width=1\textwidth]{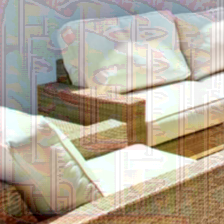}\\	
		\includegraphics[width=1\textwidth]{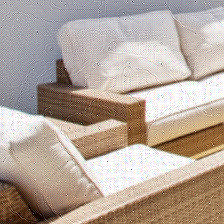}\\
		\ \\
		\includegraphics[width=1\textwidth]{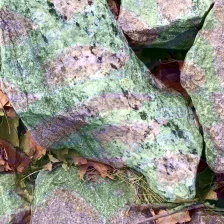}\\	
		\includegraphics[width=1\textwidth]{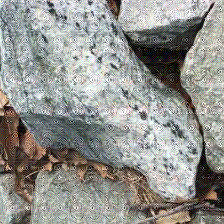}\\
	\end{minipage}
	\begin{minipage}{0.11\textwidth}
		\centering
		\small
		MuLTER~\cite{hu2019multi}
		\includegraphics[width=1\textwidth]{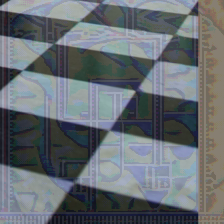}\\	
		\includegraphics[width=1\textwidth]{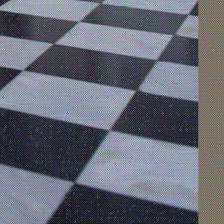}\\
		\ \\
		\includegraphics[width=1\textwidth]{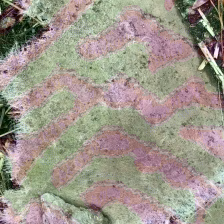}\\	
		\includegraphics[width=1\textwidth]{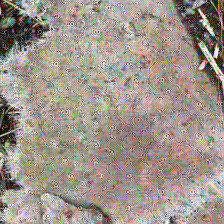}\\
	\end{minipage}
	\caption{The perturbed images generated by attacking the defended models on MINC (top two rows) and GTOS (bottom two rows). We provide visual results of adversarially perturbed images that are generated by GAP-tar~\cite{poursaeed2018generative} (first and third rows) and by our FTGAP (second and fourth rows).}
	\label{fig:viscompuat}
\end{figure}

\begin{figure*}[tb]
	\centering
	\begin{minipage}{0.10\textwidth}
		\centering
		\small
		no attack\ \\ \ \\ \ \\ \ \\ \ \\ \ \\ \ \\
		FTGAP\\
		with JND\\ \ \\ \ \\ \ \\ \ \\ \ \\ \ \\ \tiny{\ }\\
		\small
		FTGAP\\
		without JND\\ 
	\end{minipage}
	\begin{minipage}{0.13\textwidth}
		\centering
		\small
		MINC~\cite{bell2015material}
		\includegraphics[width=1\textwidth]{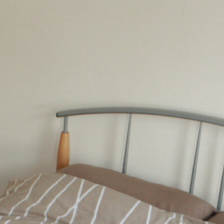}\\
		\ \\
		\tiny{\ }\\
		\includegraphics[width=1\textwidth]{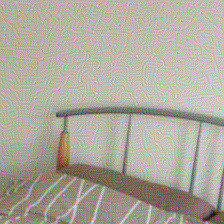}\\
		\small
		STSIM: 0.8011\\
		FR: 93.6\\
		\tiny{\ }\\
		\includegraphics[width=1\textwidth]{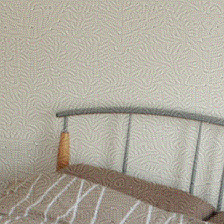}\\
		\small
		STSIM: \textbf{0.8051}\\
		FR: 93.6\\
	\end{minipage}
	\begin{minipage}{0.13\textwidth}
		\centering
		\small
		GTOS~\cite{xue2017differential, xue2018deep}
		\includegraphics[width=1\textwidth]{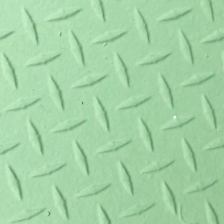}\\
		\ \\
		\tiny{\ }\\
		\includegraphics[width=1\textwidth]{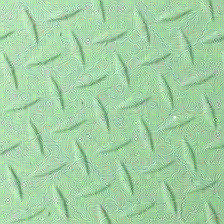}\\
		\small
		STSIM: \textbf{0.8291}\\
		FR: 73.0\\
		\tiny{\ }\\
		\includegraphics[width=1\textwidth]{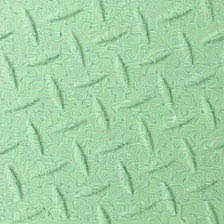}\\
		\small
		STSIM: 0.7920\\
		FR: 72.6\\
	\end{minipage}
	\begin{minipage}{0.13\textwidth}
		\centering
		\small
		DTD~\cite{cimpoi2014describing}
		\includegraphics[width=1\textwidth]{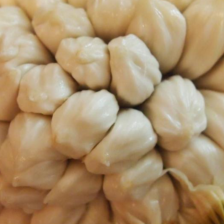}\\
		\ \\
		\tiny{\ }\\
		\includegraphics[width=1\textwidth]{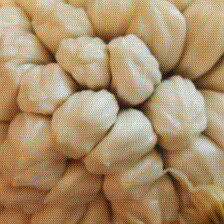}\\
		\small
		STSIM: \textbf{0.8898}\\
		FR: 78.4\\
		\tiny{\ }\\
		\includegraphics[width=1\textwidth]{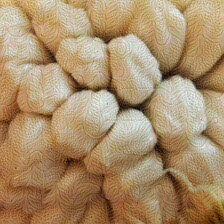}\\
		\small
		STSIM: 0.8601\\
		FR: 79.0\\
	\end{minipage}
	\begin{minipage}{0.13\textwidth}
		\centering
		\small
		4DLF~\cite{wang20164d}
		\includegraphics[width=1\textwidth]{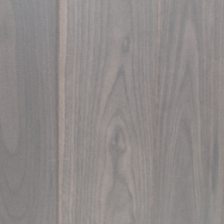}\\
		\ \\
		\tiny{\ }\\
		\includegraphics[width=1\textwidth]{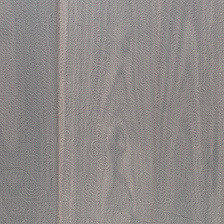}\\
		\small
		STSIM: \textbf{0.8223}\\
		FR: 89.2\\
		\tiny{\ }\\
		\includegraphics[width=1\textwidth]{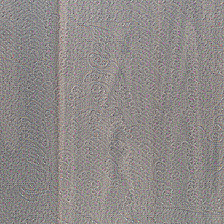}\\
		\small
		STSIM: 0.7889\\
		FR: 88.9\\
	\end{minipage}
	\begin{minipage}{0.13\textwidth}
		\centering
		\small
		FMD~\cite{sharan2013recognizing}
		\includegraphics[width=1\textwidth]{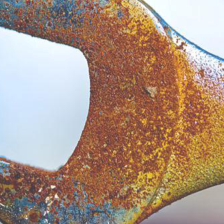}\\
		\ \\
		\tiny{\ }\\
		\includegraphics[width=1\textwidth]{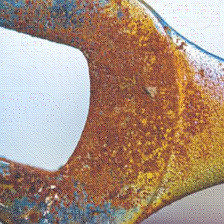}\\
		\small
		STSIM: \textbf{0.9186}\\
		FR: 92.0\\
		\tiny{\ }\\
		\includegraphics[width=1\textwidth]{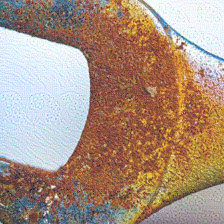}\\
		\small
		STSIM: 0.9075\\
		FR: 92.0\\
	\end{minipage}
	\begin{minipage}{0.13\textwidth}
		\centering
		\small
		KTH~\cite{caputo2005class}
		\includegraphics[width=1\textwidth]{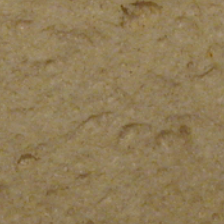}\\
		\ \\
		\tiny{\ }\\
		\includegraphics[width=1\textwidth]{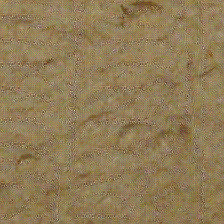}\\
		\small
		STSIM: \textbf{0.8409}\\
		FR: 73.6\\
		\tiny{\ }\\
		\includegraphics[width=1\textwidth]{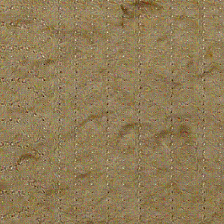}\\
		\small
		STSIM: 0.8293\\
		FR: 72.6\\
	\end{minipage}
	\caption{Comparison between FTGAP with and without JND thresholds on various datasets. The similarity index of each perturbed image in the second \& third rows with the corresponding clean image in the first row, as well as the resulting fooling rate (FR) on each dataset, are given below each perturbed image. Bold number indicates better similarity index.}
	\label{fig:jnd}
\end{figure*}

\begin{table*}
	\centering
	\caption{Cross-model fooling rates on the attacked models listed in the first row by using the FTGAP perturbations computed on the models in the first column.}
	\label{cmfr}
	
	\resizebox{0.65\textwidth}{!}{
		\begin{tabular}[t]{|c|cccc|c|cccc|c|}
			\hline
			&ResNet&DeepTEN&DEP&MuLTER&mean&ResNet&DeepTEN&DEP&MuLTER&mean\\
			\cline{2-11}
			&\multicolumn{5}{c|}{MINC}&\multicolumn{5}{c|}{GTOS}\\
			\hline
			ResNet&93.6&74.5&75.8&73.6&79.4&73.0&59.1&55.4&70.0&64.4\\
			DeepTEN&87.4&93.4&86.3&83.0&87.5&69.0&70.1&55.5&67.6&65.6\\
			DEP&85.4&84.5&94.3&82.6&86.7&65.0&69.0&75.2&69.3&69.6\\
			MuLTER&87.7&85.5&89.1&94.6&89.2&76.4&70.8&67.2&79.0&73.4\\
			\hline
			mean&88.5&84.5&86.4&83.5&85.7&70.9&67.3&63.3&71.5&68.2\\
			\hline
			&\multicolumn{5}{c|}{DTD}&\multicolumn{5}{c|}{4DLF}\\
			\hline
			ResNet&78.4&83.1&87.5&81.7&82.7&89.2&85.3&76.1&80.6&82.8\\
			DeepTEN&49.0&86.6&87.1&83.0&76.4&43.9&88.9&66.7&90.6&72.5\\
			DEP&53.0&82.5&88.8&80.4&76.2&44.7&89.4&90.0&90.3&78.6\\
			MuLTER&51.2&86.2&87.7&88.6&78.4&45.0&83.9&56.9&90.0&69.0\\
			\hline
			mean&57.9&84.6&87.8&83.4&78.4&55.7&86.9&72.4&87.9&75.7\\
			\hline
			&\multicolumn{5}{c|}{FMD}&\multicolumn{5}{c|}{KTH}\\
			\hline
			ResNet&92.0&77.0&81.0&68.0&79.5&73.6&71.9&63.5&72.6&70.4\\
			DeepTEN&34.0&94.0&80.0&83.0&72.8&9.4&82.1&82.5&79.0&63.3\\
			DEP&45.0&93.0&91.0&83.0&78.0&7.7&62.3&76.2&63.1&52.3\\
			MuLTER&36.0&90.0&85.0&86.0&74.3&9.2&77.2&78.2&85.6&62.6\\
			\hline
			mean&51.8&88.5&84.3&80.0&76.1&25.0&73.4&75.1&75.1&62.1\\
			\hline
	\end{tabular}}
\end{table*}

\begin{table*}
	\centering
	\caption{Comparisons of resulting fooling rates for baseline attack methods and their variants using the proposed frequency-tuned attack framework: sPGD (baseline)~\cite{mummadi2019defending, shafahi2020universal} vs. FT-sPGD (frequency-tuned); UPGD (baseline)~\cite{9191288} vs. FT-UPGD (frequency-tuned). Bold number indicates better performance on the corresponding model in each column within the same dataset.}
	\label{fralt}
	
	\resizebox{0.65\textwidth}{!}{
		\begin{tabular}[t]{|c|cccc|c|cccc|c|}
			\hline
			&ResNet&DeepTEN&DEP&MuLTER&mean&ResNet&DeepTEN&DEP&MuLTER&mean\\
			\cline{2-11}
			&\multicolumn{5}{c|}{MINC}&\multicolumn{5}{c|}{GTOS}\\
			\hline
			sPGD&93.8&\textbf{94.3}&93.1&93.4&93.7&61.5&74.9&70.0&76.3&70.7\\
			FT-sPGD&\textbf{94.5}&92.5&\textbf{93.7}&\textbf{94.3}&\textbf{93.8}&\textbf{61.7}&\textbf{80.8}&\textbf{76.0}&\textbf{83.7}&\textbf{75.6}\\
			\hline
			UPGD&93.4&\textbf{93.7}&93.1&93.7&93.5&78.0&77.5&72.4&77.8&76.4\\
			FT-UPGD&\textbf{93.6}&93.3&\textbf{94.4}&\textbf{94.2}&\textbf{93.9}&\textbf{81.1}&\textbf{82.3}&\textbf{76.1}&\textbf{83.9}&\textbf{80.9}\\
			\hline
			&\multicolumn{5}{c|}{DTD}&\multicolumn{5}{c|}{4DLF}\\
			\hline
			sPGD&70.9&83.9&78.8&84.5&79.5&86.4&84.4&85.6&81.7&84.5\\
			FT-sPGD&\textbf{79.3}&\textbf{87.9}&\textbf{86.6}&\textbf{85.8}&\textbf{84.9}&\textbf{89.4}&\textbf{86.7}&\textbf{90.6}&\textbf{92.8}&\textbf{89.9}\\
			\hline
			UPGD&71.8&82.6&80.5&79.9&78.7&88.3&88.3&86.1&81.9&86.2\\
			FT-UPGD&\textbf{77.0}&\textbf{88.4}&\textbf{90.4}&\textbf{88.8}&\textbf{86.2}&\textbf{89.2}&\textbf{94.7}&\textbf{92.2}&\textbf{96.1}&\textbf{93.1}\\
			\hline
			&\multicolumn{5}{c|}{FMD}&\multicolumn{5}{c|}{KTH}\\
			\hline
			sPGD&89.0&93.0&87.0&89.0&89.5&69.2&79.9&72.8&80.6&75.6\\
			FT-sPGD&\textbf{92.0}&\textbf{94.0}&\textbf{91.0}&\textbf{93.0}&\textbf{92.5}&\textbf{77.9}&\textbf{84.7}&\textbf{77.8}&\textbf{86.8}&\textbf{81.8}\\
			\hline
			UPGD&79.0&75.0&86.0&69.0&77.3&85.7&78.5&74.3&75.2&78.4\\
			FT-UPGD&\textbf{85.0}&\textbf{88.0}&\textbf{88.0}&\textbf{86.0}&\textbf{86.8}&\textbf{90.0}&\textbf{80.1}&\textbf{77.3}&\textbf{81.5}&\textbf{82.2}\\
			\hline
	\end{tabular}}
\end{table*}

\subsection{Hyperparameters for FTGAP}

For our FTGAP method, we mainly have three hyperparameters $\lambda_l$, $\lambda_h$ and $f_c$. As listed in Table~\ref{hp}\footnote[3]{For the datasets with an asterisk, we use different $\lambda_l$ and $\lambda_h$ only for ResNet. DTD: $\lambda_l = 0$ and $\lambda_h = 2$; 4DLF: $\lambda_l = 0$ and $\lambda_h = 3$; KTH: $\lambda_l = 1$ and $\lambda_h = 3$.}, we set the hyperparameters empirically to balance effectiveness and imperceptibility. An ablation study for hyperparameters is conducted in Appendix~\ref{sec:as}.

\section{Cross-Model Generalization}

Cross-model generalization assesses the transferability of the attack to other DNN models which are not used for computing the adversarial perturbation. We report the cross-model fooling rates of the perturbations that are computed by our FTGAP method on different models and datasets in Table~\ref{cmfr}. As shown in Table~\ref{cmfr}, the perturbations by our FTGAP method can still result in substantial fooling rates on unseen models for all the considered datasets.

\section{Attacking against Defended Models}

We also use GAP-tar~\cite{poursaeed2018generative} and our FTGAP to attack the models defended by universal adverarial training~\cite{shafahi2020universal} on MINC and GTOS. As shown in Table~\ref{robustuat}, our FTGAP outperforms the baseline method GAP-tar with more than 10\% drops of average top-1 accuracy on both datasets. Particularly, our FTGAP method can invalidate the defense on MINC by reducing the top-1 accuracy under 6\%, which is nearly the same as the attack performance on the undefended models. Some visual examples of perturbed images are given in Figure~\ref{fig:viscompuat}.

\section{Effects of Training Data Size}
\label{sec:trsize_a}

\begin{figure}[t]
	\centering
	\includegraphics[width=0.4\textwidth]{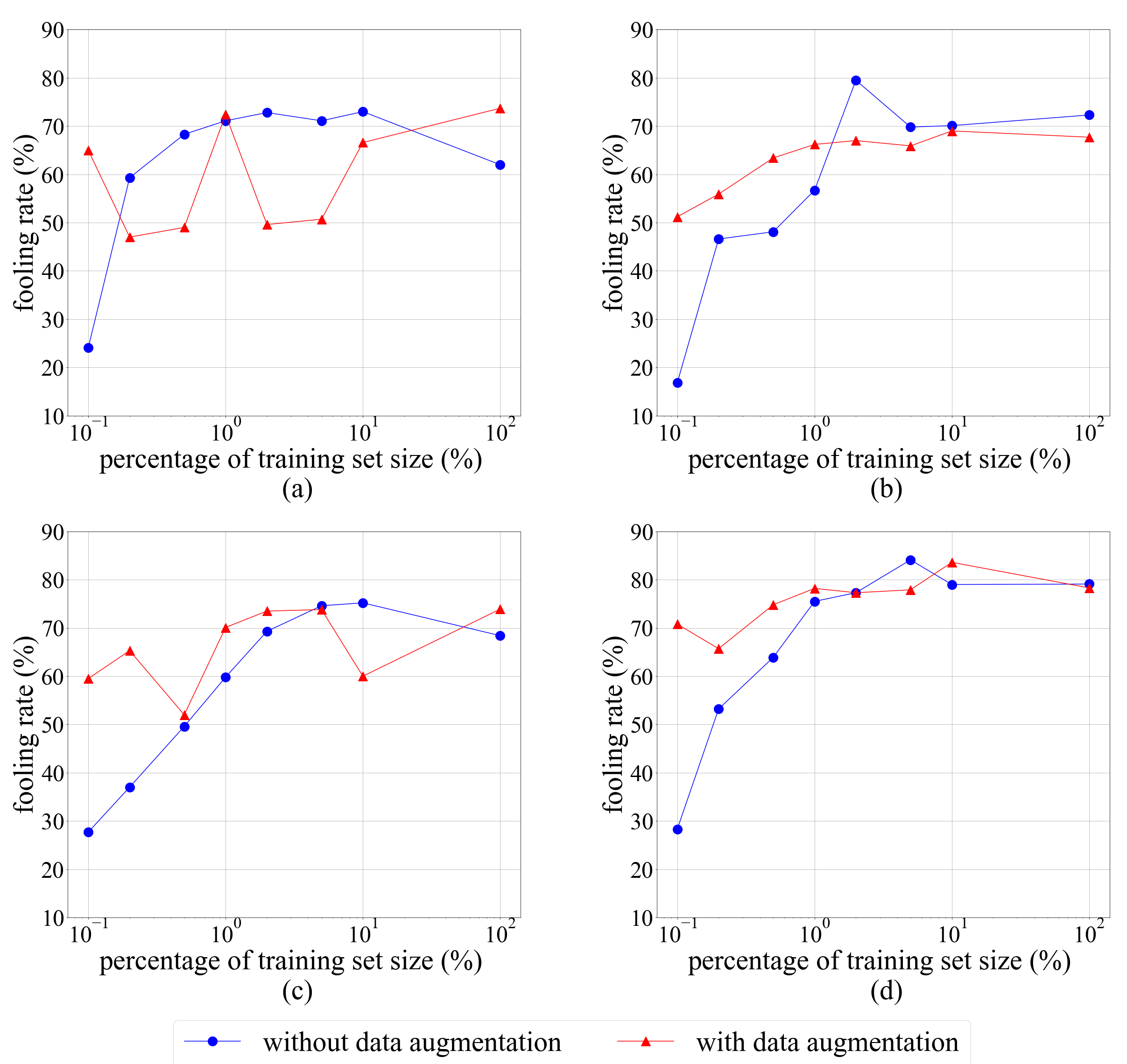}
	\caption{Fooling rate in function of the training set size (expressed as a percentage of the full training set size) using the GTOS dataset and various DNN models ((a) ResNet~\cite{he2016deep}, (b) DeepTEN~\cite{zhang2017deep}, (c) DEP~\cite{xue2018deep} and (d) MuLTER~\cite{hu2019multi}) where the training set is the set used to compute the universal perturbations.}
	\label{fig:trsize_a}
\end{figure}

Figure~\ref{fig:trsize_a} shows the effects of the training data size on the universal perturbation effectiveness for our FTGAP method using the GTOS dataset, which includes 93,945 training images. It can be seen that, despite the fooling rate fluctuations as the training set size is reduced, the proposed FTGAP with data augmentation can produce much higher fooling rates as compared to that without data augmentation even when only 0.1\% of all training images are used to compute the perturbation. The data augmentation is performed as in Section~\ref{sec:trsize} of the main paper.

\section{Ablation Study}
\label{sec:as}

\begin{figure*}[tb]
	\centering
	\begin{minipage}{0.16\textwidth}
		\centering
		\small
		no attack\\
		\includegraphics[width=1\textwidth]{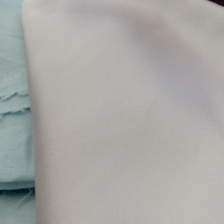}\\
		\ \\ \scriptsize{\ \\}
	\end{minipage}
	\begin{minipage}{0.08\textwidth}
		\flushright
		\small
		\textcolor[rgb]{0,0.7,0}{$\lambda_l = 0$}\ \\ \ \\ \ \\ \ \\ \ \\ \ \\ \ \\ \ \\ \ \\ \ \\
		$\lambda_l = 1$\ \\ \ \\ \ \\ \ \\ \ \\ \ \\ \ \\ \ \\ \ \\ \ \\
		$\lambda_l = 2$\ \\ \ \\
	\end{minipage}
	\begin{minipage}{0.16\textwidth}
		\centering
		\small
		$\lambda_h = 1$
		\includegraphics[width=1\textwidth]{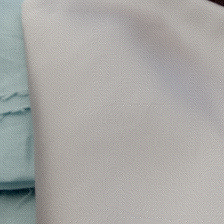}\\	
		\small{STSIM: 0.8516}\\
		FR: 58.1\\
		\tiny{\ }\\
		\includegraphics[width=1\textwidth]{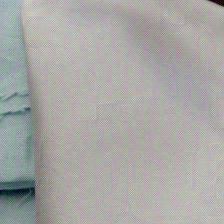}\\	
		\small{STSIM: 0.7903}\\
		FR: 83.9\\
		\tiny{\ }\\
		\includegraphics[width=1\textwidth]{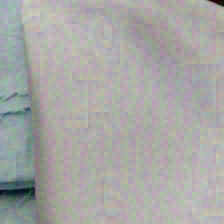}\\		
		\small{STSIM: 0.7486}\\
		FR: 86.3\\
	\end{minipage}
	\begin{minipage}{0.16\textwidth}
		\centering
		\small
		$\lambda_h = 2$
		\includegraphics[width=1\textwidth]{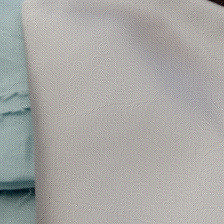}\\	
		\small{STSIM: 0.8041}\\
		FR: 83.6\\
		\tiny{\ }\\
		\includegraphics[width=1\textwidth]{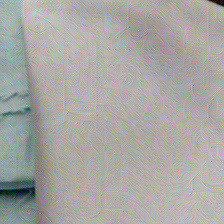}\\	
		\small{STSIM: 0.7547}\\
		FR: 87.2\\
		\tiny{\ }\\
		\includegraphics[width=1\textwidth]{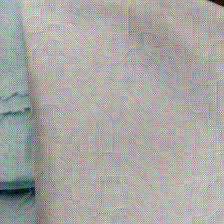}\\		
		\small{STSIM: 0.7240}\\
		FR: 89.2\\
	\end{minipage}
	\begin{minipage}{0.16\textwidth}
		\centering
		\small
		\textcolor[rgb]{0,0.7,0}{$\lambda_h = 3$}
		\includegraphics[width=1\textwidth]{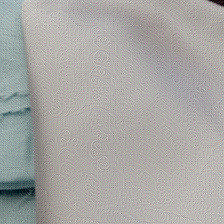}\\	
		\small{STSIM: 0.7790}\\
		FR: 89.2\\
		\tiny{\ }\\
		\includegraphics[width=1\textwidth]{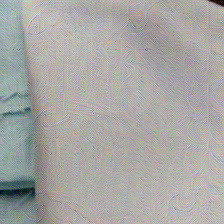}\\	
		\small{STSIM: 0.7382}\\
		FR: 89.2\\
		\tiny{\ }\\
		\includegraphics[width=1\textwidth]{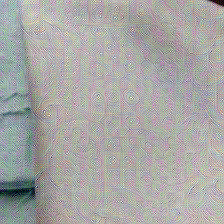}\\		
		\small{STSIM: 0.7109}\\
		FR: 90.6\\
	\end{minipage}
	\begin{minipage}{0.16\textwidth}
		\centering
		\small
		$\lambda_h = 4$
		\includegraphics[width=1\textwidth]{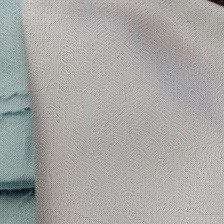}\\	
		\small{STSIM: 0.7537}\\
		FR: 91.7\\
		\tiny{\ }\\
		\includegraphics[width=1\textwidth]{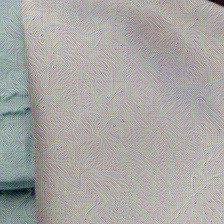}\\	
		\small{STSIM: 0.7250}\\
		FR: 90.6\\
		\tiny{\ }\\
		\includegraphics[width=1\textwidth]{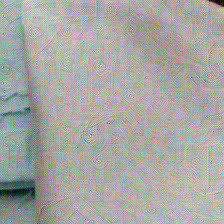}\\		
		\small{STSIM: 0.7153}\\
		FR: 91.1\\
	\end{minipage}
	\caption{Effects of $\lambda_l$ and $\lambda_h$ for the ResNet model~\cite{he2016deep} and the 4DLF dataset~\cite{wang20164d}. Under each perturbed image, the similarity index with the clean image and the fooling rate (FR) obtained by the corresponding perturbation are given. The hyperparameter settings that are used in our method for producing the results in the main paper, are shown in green.}
	\label{fig:lambda}
\end{figure*}

\begin{figure*}[tb]
	\centering
	\begin{minipage}{0.13\textwidth}
		\centering
		\small
		no attack\\
		\includegraphics[width=1\textwidth]{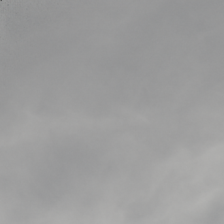}\\
		\ \\ \ \\
	\end{minipage}
	\begin{minipage}{0.13\textwidth}
		\centering
		\small
		$f_c = 0$\\
		\includegraphics[width=1\textwidth]{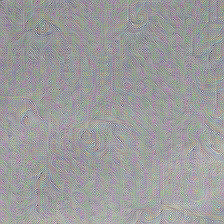}\\
		STSIM: 0.6715\\
		FR: 89.7\\
	\end{minipage}
	\begin{minipage}{0.13\textwidth}
		\centering
		\small
		$f_c = 2$\\
		\includegraphics[width=1\textwidth]{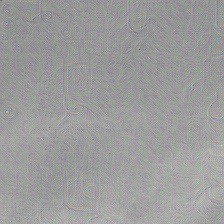}\\
		STSIM: 0.7039\\
		FR: 89.2\\
	\end{minipage}
	\begin{minipage}{0.13\textwidth}
		\centering
		\small
		\textcolor[rgb]{0,0.7,0}{$f_c = 4$}\\
		\includegraphics[width=1\textwidth]{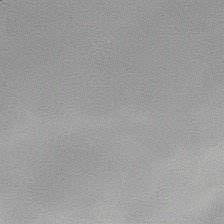}\\
		STSIM: 0.7423\\
		FR: 89.2\\
	\end{minipage}
	\begin{minipage}{0.13\textwidth}
		\centering
		\small
		$f_c = 6$\\
		\includegraphics[width=1\textwidth]{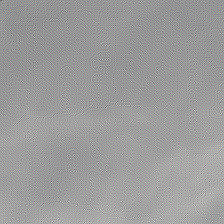}\\
		STSIM: 0.7959\\
		FR: 88.3\\
	\end{minipage}
	\begin{minipage}{0.13\textwidth}
		\centering
		\small
		$f_c = 8$\\
		\includegraphics[width=1\textwidth]{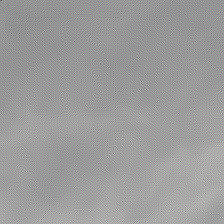}\\
		STSIM: 0.8082\\
		FR: 86.9\\
	\end{minipage}
	\begin{minipage}{0.13\textwidth}
		\centering
		\small
		$f_c = 10$\\
		\includegraphics[width=1\textwidth]{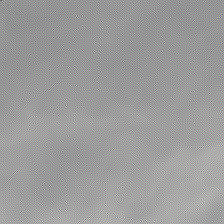}\\
		STSIM: 0.8290\\
		FR: 85.0\\
	\end{minipage}
	\caption{Effects of $f_c$ for the ResNet model~\cite{he2016deep} and the 4DLF dataset~\cite{wang20164d}. Under each perturbed image, the similarity index with the clean image and the fooling rate (FR) obtained by the corresponding perturbation are given. The hyperparameter setting that is used in our method for producing the experimental results in the main paper, is shown in green.}
	\label{fig:fc}
\end{figure*}

\textbf{JND thresholds.} To exhibit the effect of the adopted JND thresholding model on imperceptibility, we examine our FTGAP method with and without JND thresholds on the ResNet model~\cite{he2016deep}. For FTGAP without JND thresholds, we empirically find a constant to replace the JND thresholds over all frequency bands while producing a fooling rate result that is similar to FTGAP with JND thresholds, and we compare the imperceptibility of the computed and applied adversarial perturbations. According to Figure~\ref{fig:jnd}, the adopted JND-based thresholding model can help improve the imperceptibility of the generated adversarial perturbations.

\textbf{$\lambda_l$ and $\lambda_h$.} The effects of $\lambda_l$ and $\lambda_h$ for low and high frequency bands, respectively, are illustrated based on the ResNet model~\cite{he2016deep} and the 4DLF dataset~\cite{wang20164d} in Figure~\ref{fig:lambda}. From Figure~\ref{fig:lambda}, it can be seen that increasing both $\lambda_l$ and $\lambda_h$ can result in higher fooling rates but in lower similarity values (i.e., more visible perturbations). In particular, increasing $\lambda_l$ can significantly increase the perceptibility of the adversarial perturbation as compared to an increase in the value of $\lambda_h$, which is consistent with the fact that perturbations in low frequency components can be more perceived as compared to perturbations in high frequency components. Furthermore, Figure~\ref{fig:lambda} shows that increasing $\lambda_h$ can result in a higher fooling rate and in a less perceptible perturbation as compared to $\lambda_l$.

\textbf{$f_c$.} We illustrate the effects of the cut-off frequency $f_c$ based on the ResNet model~\cite{he2016deep} and the 4DLF dataset~\cite{wang20164d} in Figure~\ref{fig:fc} for $\lambda_l=0$ and $\lambda_h=3$. A higher $f_c$ value will result in more frequency bands being treated as belonging to the low frequency region ($f<f_c$) while shrinking the high frequency region to bands with higher indices (corresponding to $f>f_c$). The higher $f_c$ will limit the computed perturbation to the higher frequency bands in the DCT domain. According to Figure~\ref{fig:fc}, the fooling rate becomes saturated when we reduce $f_c$ to be less than 4 cycles/degree.

\section{Visual Comparisons of More Examples}

\begin{figure*}[tb]
	\centering
	\begin{minipage}{0.11\textwidth}
		\centering
		\small
		MINC~\cite{bell2015material}\\
		\&\\
		DeepTEN~\cite{zhang2017deep}\\ \ \\ \ \\ \ \\ \ \\ \ \\ \ \\
		GTOS~\cite{xue2017differential, xue2018deep}\\
		\&\\
		ResNet~\cite{he2016deep}\\ \ \\ \ \\ \ \\ \ \\ \ \\ \tiny{\ \\}
		\small
		DTD~\cite{cimpoi2014describing}\\
		\&\\
		MuLTER~\cite{hu2019multi}\\ \ \\ \ \\ \ \\ \ \\ \ \\ \ \\
		4DLF~\cite{wang20164d}\\
		\&\\
		DEP~\cite{xue2018deep}\\ \ \\ \ \\ \ \\ \ \\ \ \\ \tiny{\ \\}
		\small
		FMD~\cite{sharan2013recognizing}\\
		\&\\
		ResNet~\cite{he2016deep}\\ \ \\ \ \\ \ \\ \ \\ \ \\ \ \\
		KTH~\cite{caputo2005class}\\
		\&\\
		MuLTER~\cite{hu2019multi}
	\end{minipage}
	\begin{minipage}{0.16\textwidth}
		\small
		\centering
		no attack
		\includegraphics[width=1\textwidth]{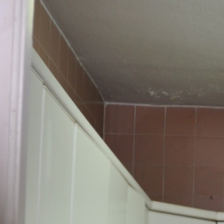}\\	
		\small{\ }\\
		\tiny{\ }\\
		\includegraphics[width=1\textwidth]{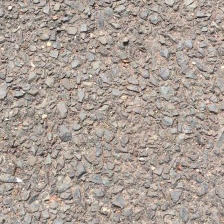}\\	
		\small{\ }\\
		\tiny{\ }\\
		\includegraphics[width=1\textwidth]{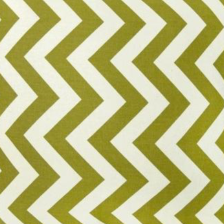}\\		
		\small{\ }\\
		\tiny{\ }\\
		\includegraphics[width=1\textwidth]{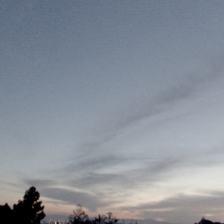}\\		
		\small{\ }\\
		\tiny{\ }\\
		\includegraphics[width=1\textwidth]{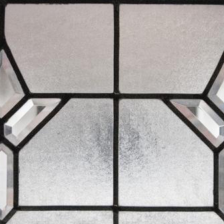}\\		
		\small{\ }\\
		\tiny{\ }\\
		\includegraphics[width=1\textwidth]{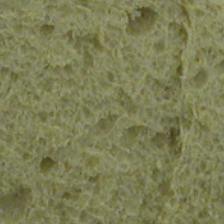}\\
		\small{\ }\\
	\end{minipage}
	\begin{minipage}{0.16\textwidth}
		\centering
		\small
		GAP-tar~\cite{poursaeed2018generative}
		\includegraphics[width=1\textwidth]{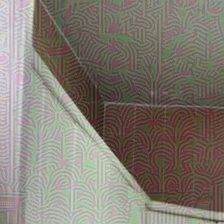}\\	
		\small{STSIM: 0.7934}\\
		\tiny{\ }\\
		\includegraphics[width=1\textwidth]{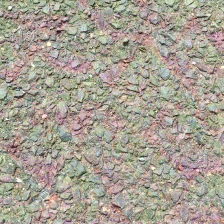}\\	
		\small{STSIM: 0.9410}\\
		\tiny{\ }\\
		\includegraphics[width=1\textwidth]{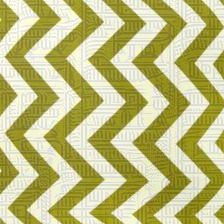}\\		
		\small{STSIM: 0.8788}\\
		\tiny{\ }\\
		\includegraphics[width=1\textwidth]{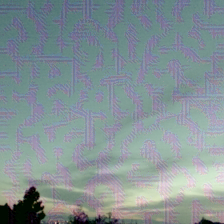}\\		
		\small{STSIM: 0.8083}\\
		\tiny{\ }\\
		\includegraphics[width=1\textwidth]{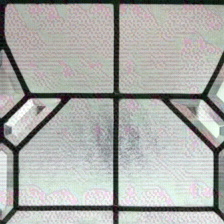}\\		
		\small{STSIM: 0.9135}\\
		\tiny{\ }\\
		\includegraphics[width=1\textwidth]{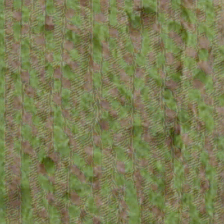}\\
		\small{STSIM: 0.9112}\\
	\end{minipage}
	\begin{minipage}{0.16\textwidth}
		\centering
		sPGD~\cite{mummadi2019defending, shafahi2020universal}
		\includegraphics[width=1\textwidth]{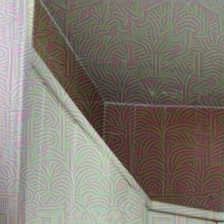}\\	
		\small{STSIM: 0.8032}\\
		\tiny{\ }\\
		\includegraphics[width=1\textwidth]{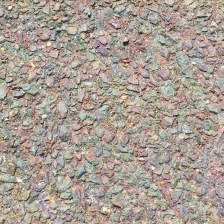}\\	
		\small{STSIM: 0.9319}\\
		\tiny{\ }\\
		\includegraphics[width=1\textwidth]{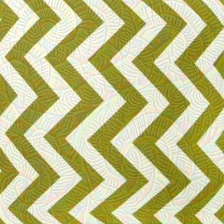}\\		
		\small{STSIM: 0.9030}\\
		\tiny{\ }\\
		\includegraphics[width=1\textwidth]{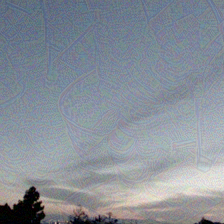}\\		
		\small{STSIM: 0.8082}\\
		\tiny{\ }\\
		\includegraphics[width=1\textwidth]{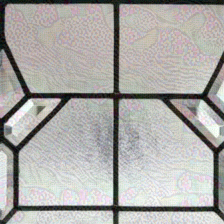}\\		
		\small{STSIM: 0.8982}\\
		\tiny{\ }\\
		\includegraphics[width=1\textwidth]{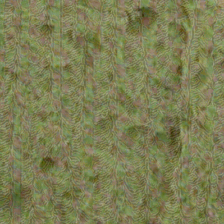}\\
		\small{STSIM: 0.8901}\\
	\end{minipage}
	\begin{minipage}{0.16\textwidth}
		\centering
		UPGD~\cite{9191288}
		\includegraphics[width=1\textwidth]{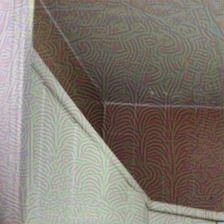}\\	
		\small{STSIM: 0.7954}\\
		\tiny{\ }\\
		\includegraphics[width=1\textwidth]{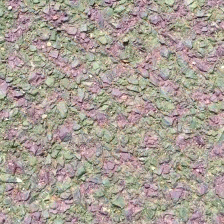}\\	
		\small{STSIM: \textbf{0.9565}}\\
		\tiny{\ }\\
		\includegraphics[width=1\textwidth]{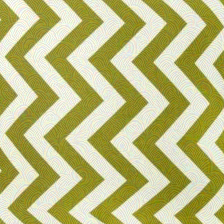}\\		
		\small{STSIM: 0.9374}\\
		\tiny{\ }\\
		\includegraphics[width=1\textwidth]{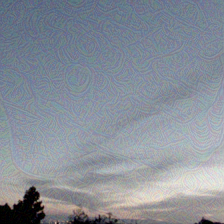}\\		
		\small{STSIM: 0.8094}\\
		\tiny{\ }\\
		\includegraphics[width=1\textwidth]{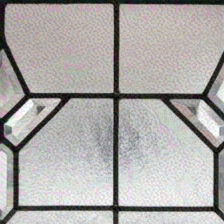}\\		
		\small{STSIM: 0.9081}\\
		\tiny{\ }\\
		\includegraphics[width=1\textwidth]{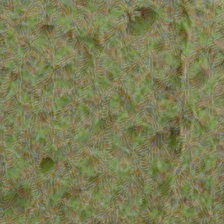}\\
		\small{STSIM: 0.9022}\\
	\end{minipage}
	\begin{minipage}{0.16\textwidth}
		\centering
		FTGAP (ours)
		\includegraphics[width=1\textwidth]{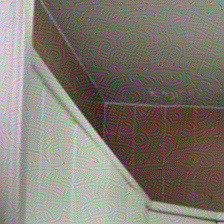}\\	
		\small{STSIM: \textbf{0.8082}}\\
		\tiny{\ }\\
		\includegraphics[width=1\textwidth]{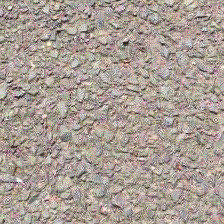}\\	
		\small{STSIM: 0.9456}\\
		\tiny{\ }\\
		\includegraphics[width=1\textwidth]{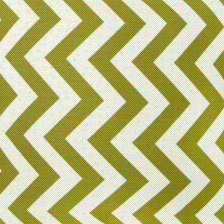}\\		
		\small{STSIM: \textbf{0.9403}}\\
		\tiny{\ }\\
		\includegraphics[width=1\textwidth]{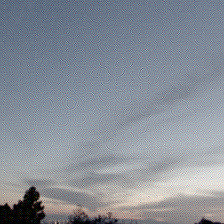}\\		
		\small{STSIM: \textbf{0.8458}}\\
		\tiny{\ }\\
		\includegraphics[width=1\textwidth]{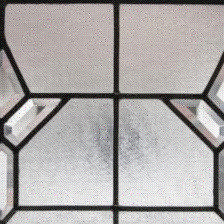}\\		
		\small{STSIM: \textbf{0.9175}}\\
		\tiny{\ }\\
		\includegraphics[width=1\textwidth]{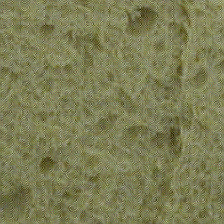}\\
		\small{STSIM: \textbf{0.9273}}\\
	\end{minipage}
	\caption{Visual comparison of different universal attack methods. Each row shows the perturbed images that result from applying on the clean image (leftmost image), universal perturbations that are generated by the considered universal attack methods for different DNN models and datasets (shown in leftmost column). The structural texture similarity (STSIM) index is given below each perturbed image. Bold number indicates best performance for the model \& dataset in each row.}
	\label{fig:viscomp1}
\end{figure*}
\begin{figure*}[tb]
	\centering
	\begin{minipage}{0.11\textwidth}
		\centering
		\small
		MINC~\cite{bell2015material}\\
		\&\\
		DeepTEN~\cite{zhang2017deep}\\ \ \\ \ \\ \ \\ \ \\ \ \\ \ \\
		GTOS~\cite{xue2017differential, xue2018deep}\\
		\&\\
		ResNet~\cite{he2016deep}\\ \ \\ \ \\ \ \\ \ \\ \ \\ \tiny{\ \\}
		\small
		DTD~\cite{cimpoi2014describing}\\
		\&\\
		MuLTER~\cite{hu2019multi}\\ \ \\ \ \\ \ \\ \ \\ \ \\ \ \\
		4DLF~\cite{wang20164d}\\
		\&\\
		DEP~\cite{xue2018deep}\\ \ \\ \ \\ \ \\ \ \\ \ \\ \tiny{\ \\}
		\small
		FMD~\cite{sharan2013recognizing}\\
		\&\\
		ResNet~\cite{he2016deep}\\ \ \\ \ \\ \ \\ \ \\ \ \\ \ \\
		KTH~\cite{caputo2005class}\\
		\&\\
		MuLTER~\cite{hu2019multi}
	\end{minipage}
	\begin{minipage}{0.16\textwidth}
		\centering
		no attack
		\includegraphics[width=1\textwidth]{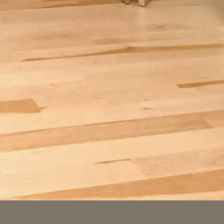}\\	
		\small{\ }\\
		\tiny{\ }\\
		\includegraphics[width=1\textwidth]{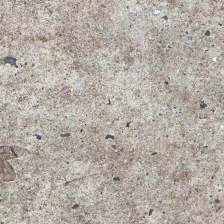}\\	
		\small{\ }\\
		\tiny{\ }\\
		\includegraphics[width=1\textwidth]{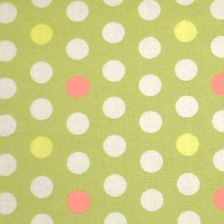}\\		
		\small{\ }\\
		\tiny{\ }\\
		\includegraphics[width=1\textwidth]{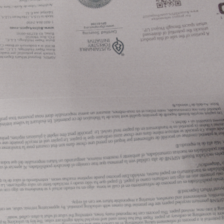}\\		
		\small{\ }\\
		\tiny{\ }\\
		\includegraphics[width=1\textwidth]{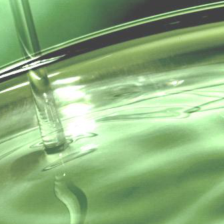}\\		
		\small{\ }\\
		\tiny{\ }\\
		\includegraphics[width=1\textwidth]{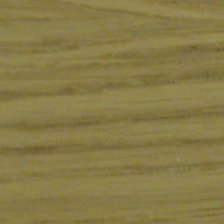}\\
		\small{\ }\\
	\end{minipage}
	\begin{minipage}{0.16\textwidth}
		\centering
		GAP-tar~\cite{poursaeed2018generative}
		\includegraphics[width=1\textwidth]{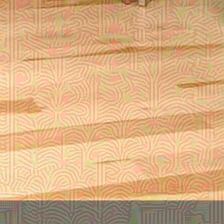}\\	
		\small{STSIM: 0.7956}\\
		\tiny{\ }\\
		\includegraphics[width=1\textwidth]{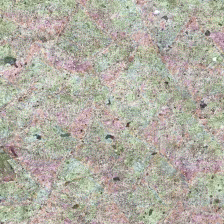}\\	
		\small{STSIM: 0.9177}\\
		\tiny{\ }\\
		\includegraphics[width=1\textwidth]{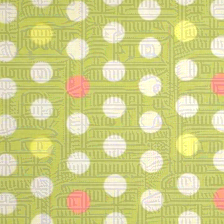}\\		
		\small{STSIM: 0.8225}\\
		\tiny{\ }\\
		\includegraphics[width=1\textwidth]{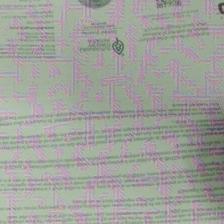}\\		
		\small{STSIM: 0.8607}\\
		\tiny{\ }\\
		\includegraphics[width=1\textwidth]{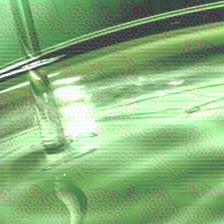}\\		
		\small{STSIM: 0.8715}\\
		\tiny{\ }\\
		\includegraphics[width=1\textwidth]{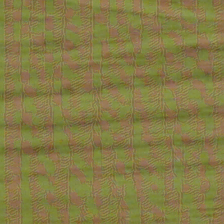}\\
		\small{STSIM: 0.8220}\\
	\end{minipage}
	\begin{minipage}{0.16\textwidth}
		\centering
		sPGD~\cite{mummadi2019defending, shafahi2020universal}
		\includegraphics[width=1\textwidth]{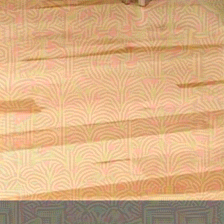}\\	
		\small{STSIM: 0.8033}\\
		\tiny{\ }\\
		\includegraphics[width=1\textwidth]{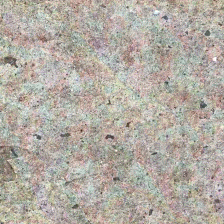}\\	
		\small{STSIM: 0.9060}\\
		\tiny{\ }\\
		\includegraphics[width=1\textwidth]{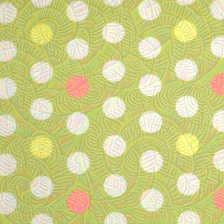}\\		
		\small{STSIM: 0.8467}\\
		\tiny{\ }\\
		\includegraphics[width=1\textwidth]{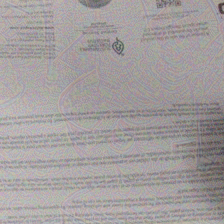}\\		
		\small{STSIM: 0.8627}\\
		\tiny{\ }\\
		\includegraphics[width=1\textwidth]{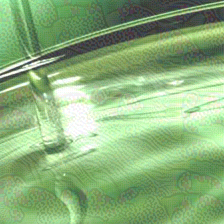}\\		
		\small{STSIM: 0.8518}\\
		\tiny{\ }\\
		\includegraphics[width=1\textwidth]{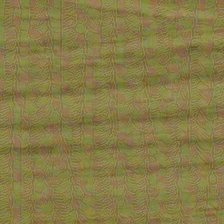}\\
		\small{STSIM: 0.7980}\\
	\end{minipage}
	\begin{minipage}{0.16\textwidth}
		\centering
		UPGD~\cite{9191288}
		\includegraphics[width=1\textwidth]{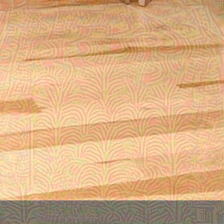}\\	
		\small{STSIM: 0.7987}\\
		\tiny{\ }\\
		\includegraphics[width=1\textwidth]{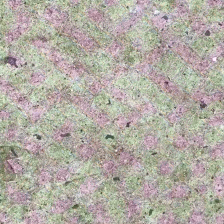}\\	
		\small{STSIM: \textbf{0.9384}}\\
		\tiny{\ }\\
		\includegraphics[width=1\textwidth]{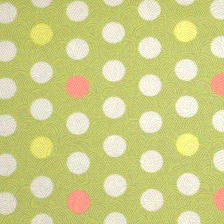}\\		
		\small{STSIM: 0.8974}\\
		\tiny{\ }\\
		\includegraphics[width=1\textwidth]{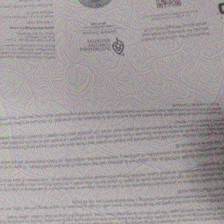}\\		
		\small{STSIM: 0.8606}\\
		\tiny{\ }\\
		\includegraphics[width=1\textwidth]{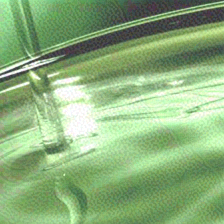}\\		
		\small{STSIM: 0.8646}\\
		\tiny{\ }\\
		\includegraphics[width=1\textwidth]{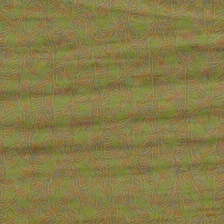}\\
		\small{STSIM: 0.8139}\\
	\end{minipage}
	\begin{minipage}{0.16\textwidth}
		\centering
		FTGAP (ours)
		\includegraphics[width=1\textwidth]{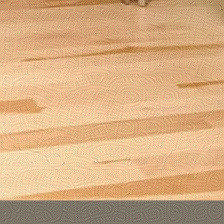}\\	
		\small{STSIM: \textbf{0.8056}}\\
		\tiny{\ }\\
		\includegraphics[width=1\textwidth]{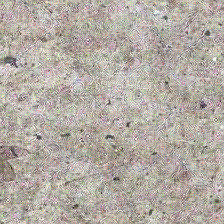}\\	
		\small{STSIM: 0.9235}\\
		\tiny{\ }\\
		\includegraphics[width=1\textwidth]{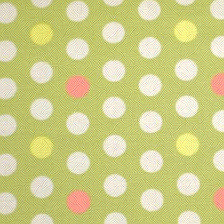}\\		
		\small{STSIM: \textbf{0.9111}}\\
		\tiny{\ }\\
		\includegraphics[width=1\textwidth]{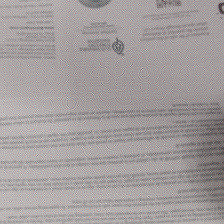}\\		
		\small{STSIM: \textbf{0.8954}}\\
		\tiny{\ }\\
		\includegraphics[width=1\textwidth]{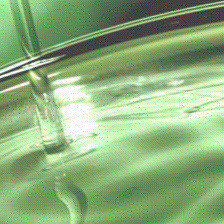}\\		
		\small{STSIM: \textbf{0.8746}}\\
		\tiny{\ }\\
		\includegraphics[width=1\textwidth]{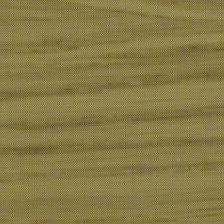}\\
		\small{STSIM: \textbf{0.8615}}\\
	\end{minipage}
	\caption{Visual comparison of different universal attack methods. Each row shows the perturbed images that result from applying on the clean image (leftmost image), universal perturbations that are generated by the considered universal attack methods for different DNN models and datasets (shown in leftmost column). The structural texture similarity (STSIM) index is given below each perturbed image. Bold number indicates best performance for the model \& dataset in each row.}
	\label{fig:viscomp2}
\end{figure*}

Additional visual comparisons of perturbed images that are generated by our FTGAP method and by existing universal attack methods are provided in Figures~\ref{fig:viscomp1} and~\ref{fig:viscomp2}.

\section{Results for Alternative Baselines}
\label{sec:alt}

\begin{figure*}[tb]
	\centering
	\begin{minipage}{0.11\textwidth}
		\centering
		\small
		MINC~\cite{bell2015material}\ \\ \ \\ \ \\ \ \\ \ \\ \ \\ \ \\ \ \\
		GTOS~\cite{xue2017differential, xue2018deep}\ \\ \ \\ \ \\ \ \\ \ \\ \ \\ \ \\ \ \\ \ \\
		DTD~\cite{cimpoi2014describing}\ \\ \ \\ \ \\ \ \\ \ \\ \ \\ \ \\ \ \\
		4DLF~\cite{wang20164d}\ \\ \ \\ \ \\ \ \\ \ \\ \ \\ \ \\ \ \\ \ \\
		FMD~\cite{sharan2013recognizing}\ \\ \ \\ \ \\ \ \\ \ \\ \ \\ \ \\ \ \\
		KTH~\cite{caputo2005class}\\ \ \\
	\end{minipage}
	\begin{minipage}{0.13\textwidth}
		\centering
		\small
		no attack
		\includegraphics[width=1\textwidth]{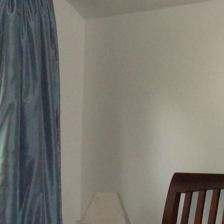}\\	
		\small{\ }\\
		\small{\ }\\
		\tiny{\ }\\
		\includegraphics[width=1\textwidth]{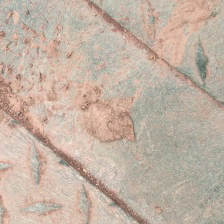}\\	
		\small{\ }\\
		\small{\ }\\
		\tiny{\ }\\
		\includegraphics[width=1\textwidth]{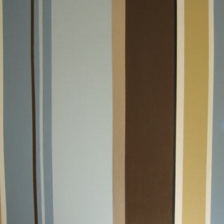}\\		
		\small{\ }\\
		\small{\ }\\
		\tiny{\ }\\
		\includegraphics[width=1\textwidth]{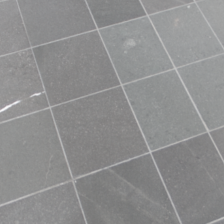}\\		
		\small{\ }\\
		\small{\ }\\
		\tiny{\ }\\
		\includegraphics[width=1\textwidth]{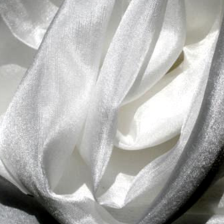}\\		
		\small{\ }\\
		\small{\ }\\
		\tiny{\ }\\
		\includegraphics[width=1\textwidth]{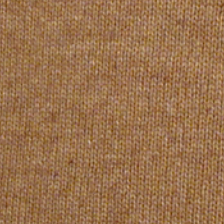}\\
		\small{\ }\\
		\small{\ }\\
	\end{minipage}
	\begin{minipage}{0.01\textwidth}
		\ \\
	\end{minipage}
	\begin{minipage}{0.13\textwidth}
		\centering
		\small
		sPGD~\cite{mummadi2019defending, shafahi2020universal}
		\includegraphics[width=1\textwidth]{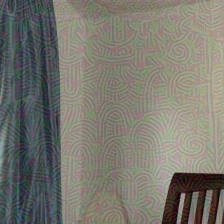}\\	
		\small{STSIM: 0.8095}\\
		\small{FR: 93.1}\\
		\tiny{\ }\\
		\includegraphics[width=1\textwidth]{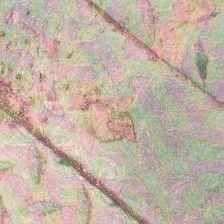}\\	
		\small{\textbf{STSIM: 0.9089}}\\
		\small{FR: 70.0}\\
		\tiny{\ }\\
		\includegraphics[width=1\textwidth]{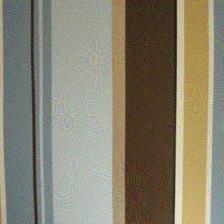}\\		
		\small{STSIM: 0.8300}\\
		\small{FR: 78.8}\\
		\tiny{\ }\\
		\includegraphics[width=1\textwidth]{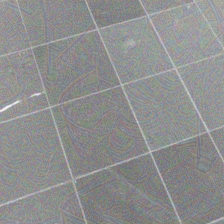}\\		
		\small{STSIM: 0.8585}\\
		\small{FR: 85.6}\\
		\tiny{\ }\\
		\includegraphics[width=1\textwidth]{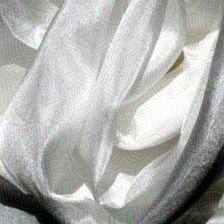}\\		
		\small{STSIM: 0.8957}\\
		\small{FR: 87.0}\\
		\tiny{\ }\\
		\includegraphics[width=1\textwidth]{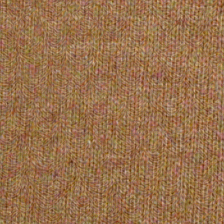}\\
		\small{STSIM: 0.9268}\\
		\small{FR: 72.8}\\
	\end{minipage}
	\begin{minipage}{0.13\textwidth}
		\centering
		\small
		FT-sPGD (ours)
		\includegraphics[width=1\textwidth]{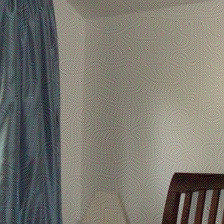}\\	
		\small{\textbf{STSIM: 0.8373}}\\
		\small{\textbf{FR: 93.7}}\\
		\tiny{\ }\\
		\includegraphics[width=1\textwidth]{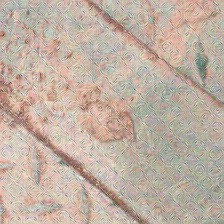}\\	
		\small{STSIM: 0.8821}\\
		\small{\textbf{FR: 76.0}}\\
		\tiny{\ }\\
		\includegraphics[width=1\textwidth]{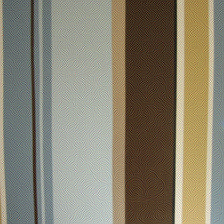}\\		
		\small{\textbf{STSIM: 0.8738}}\\
		\small{\textbf{FR: 86.6}}\\
		\tiny{\ }\\
		\includegraphics[width=1\textwidth]{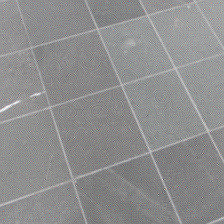}\\		
		\small{\textbf{STSIM: 0.8985}}\\
		\small{\textbf{FR: 90.6}}\\
		\tiny{\ }\\
		\includegraphics[width=1\textwidth]{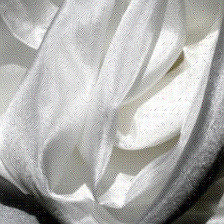}\\		
		\small{\textbf{STSIM: 0.9017}}\\
		\small{\textbf{FR: 91.0}}\\
		\tiny{\ }\\
		\includegraphics[width=1\textwidth]{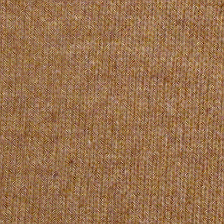}\\
		\small{\textbf{STSIM: 0.9581}}\\
		\small{\textbf{FR: 77.8}}\\
	\end{minipage}
	\begin{minipage}{0.01\textwidth}
		\ \\
	\end{minipage}
	\begin{minipage}{0.13\textwidth}
		\centering
		\small
		UPGD~\cite{9191288}
		\includegraphics[width=1\textwidth]{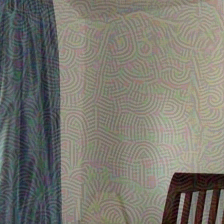}\\	
		\small{STSIM: 0.8009}\\
		\small{FR: 93.1}\\
		\tiny{\ }\\
		\includegraphics[width=1\textwidth]{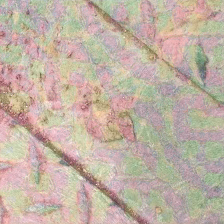}\\	
		\small{STSIM: 0.8893}\\
		\small{FR: 72.4}\\
		\tiny{\ }\\
		\includegraphics[width=1\textwidth]{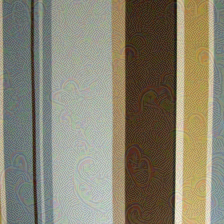}\\		
		\small{STSIM: 0.8276}\\
		\small{FR: 80.5}\\
		\tiny{\ }\\
		\includegraphics[width=1\textwidth]{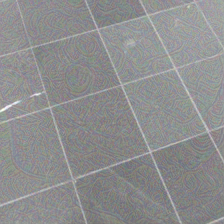}\\		
		\small{STSIM: 0.8582}\\
		\small{FR: 86.1}\\
		\tiny{\ }\\
		\includegraphics[width=1\textwidth]{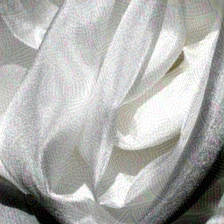}\\		
		\small{STSIM: 0.9011}\\
		\small{FR: 86.0}\\
		\tiny{\ }\\
		\includegraphics[width=1\textwidth]{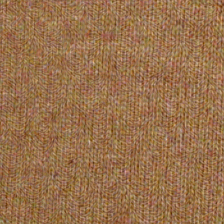}\\
		\small{STSIM: 0.9298}\\
		\small{FR: 74.3}\\
	\end{minipage}
	\begin{minipage}{0.13\textwidth}
		\centering
		\small
		FT-UPGD (ours)
		\includegraphics[width=1\textwidth]{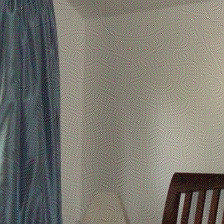}\\	
		\small{\textbf{STSIM: 0.8281}}\\
		\small{\textbf{FR: 94.4}}\\
		\tiny{\ }\\
		\includegraphics[width=1\textwidth]{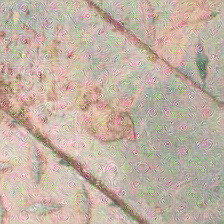}\\	
		\small{\textbf{STSIM: 0.9039}}\\
		\small{\textbf{FR: 76.1}}\\
		\tiny{\ }\\
		\includegraphics[width=1\textwidth]{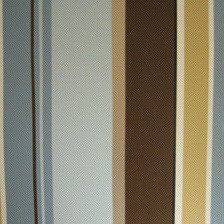}\\		
		\small{\textbf{STSIM: 0.8691}}\\
		\small{\textbf{FR: 90.4}}\\
		\tiny{\ }\\
		\includegraphics[width=1\textwidth]{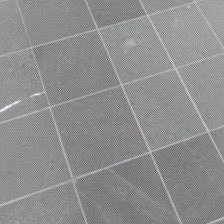}\\		
		\small{\textbf{STSIM: 0.8944}}\\
		\small{\textbf{FR: 92.2}}\\
		\tiny{\ }\\
		\includegraphics[width=1\textwidth]{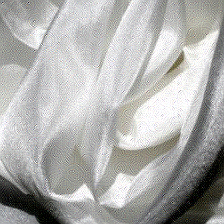}\\		
		\small{\textbf{STSIM: 0.9095}}\\
		\small{\textbf{FR: 88.0}}\\
		\tiny{\ }\\
		\includegraphics[width=1\textwidth]{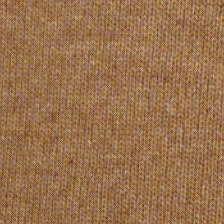}\\
		\small{\textbf{STSIM: 0.9580}}\\
		\small{\textbf{FR: 77.3}}\\
	\end{minipage}
	\caption{Visual comparison of different universal attack methods. Each row shows the perturbed images that result from applying on the clean image (leftmost image), universal perturbations that are generated by the considered universal attack methods (shown in top row) for the DEP model~\cite{xue2018deep} and different datasets (shown in leftmost column). The structural texture similarity (STSIM) index and resulting fooling rate (FR) are given below each perturbed image. Bold numbers indicate best performance for each image in a row when comparing each baseline attack method with its variant using our proposed frequency-tuned (FT) attack framework.}
	\label{fig:viscompalt}
\end{figure*}

Our frequency-tuned method can be easily adapted to improve the performance of other baseline methods, such as sPGD~\cite{mummadi2019defending, shafahi2020universal} and UPGD~\cite{9191288}. We evaluate our proposed frequency-tuned attack framework based on sPGD and UPGD, which we refer to as FT-sPGD and FT-UPGD, respectively. The obtained performance results (Table~\ref{fralt}, Figure~\ref{fig:viscompalt}) show that our proposed frequency-tuned approach can enhance the adversarial attack strengths of the baseline methods in terms of white-box fooling rates on almost all the datasets and DNN models as shown in Table~\ref{fralt}, with a reduced perceptibility as illustrated in Figure~\ref{fig:viscompalt}.

\end{document}